\pdfoutput=1

\documentclass[11pt]{article}

\usepackage[final]{acl}

\usepackage{times}
\usepackage{latexsym}

\usepackage[T1]{fontenc}

\usepackage[utf8]{inputenc}

\usepackage{microtype}

\usepackage{inconsolata}

\usepackage{graphicx}

\PassOptionsToPackage{table}{xcolor}
\usepackage{xcolor}
\usepackage{colortbl}
\usepackage{booktabs}
\usepackage{graphicx}
\usepackage{tcolorbox}
\usepackage{mdframed}

\usepackage{url}
\usepackage{listings}

\usepackage{amssymb}
\usepackage{multirow}

\usepackage{amsmath}
\usepackage{enumitem}

\newtheorem{theorem}{Theorem}[section]

\title{\includegraphics[height=0.5cm]{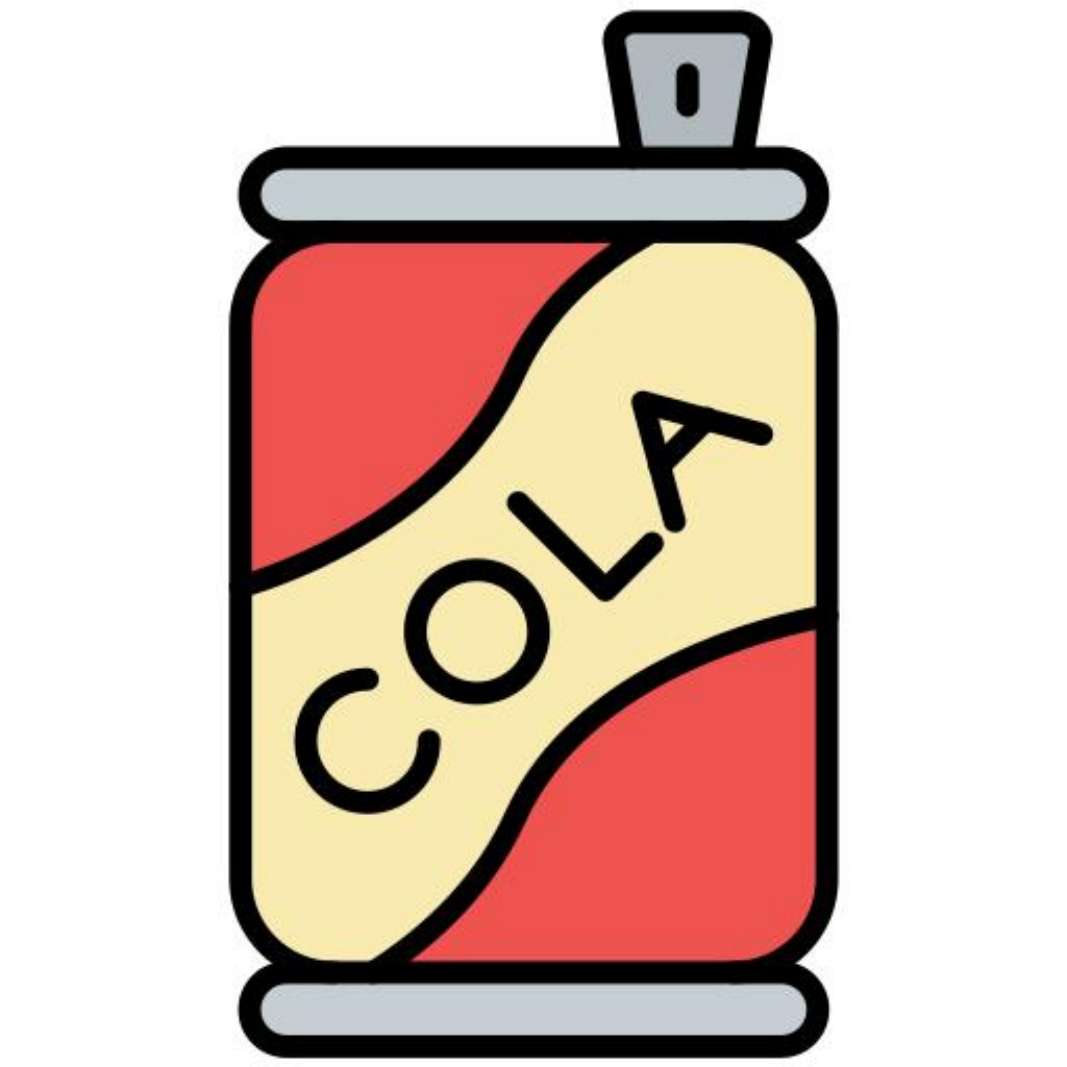}CoLA: \underline{Co}llaborative \underline{L}ow-Rank \underline{A}daptation}

\author{Yiyun Zhou,\quad Chang Yao,\quad Jingyuan Chen\thanks{Corresponding author.} \\
    Zhejiang University \\
    {\small\{yiyunzhou, changy, jingyuanchen\}@zju.edu.cn}
  }

\begin{document}
\maketitle

\newtcolorbox{myboxi}[1][Default Title]{
  title=#1,
  colback=red!5,
  colbacktitle=red!5,
  coltitle=black,
  fonttitle=\bfseries,
  bottomrule=0pt,
  toprule=0pt,
  leftrule=1.5pt,
  rightrule=1.5pt,
  titlerule=0.5pt,
  arc=0pt,
  outer arc=0pt,
  colframe=red!50,
}

\newmdenv[
  backgroundcolor=orange!10,
  skipabove=1em,
  skipbelow=0em,
  leftline=true,
  topline=false,
  bottomline=false,
  rightline=false,
  linecolor=orange!70,
  linewidth=3pt
]{githubquote}

\begin{abstract}
\label{sec:abstract}

The scaling law of Large Language Models (LLMs) reveals a power-law relationship, showing diminishing return on performance as model scale increases. While training LLMs from scratch is resource-intensive, fine-tuning a pre-trained model for specific tasks has become a practical alternative. Full fine-tuning (FFT) achieves strong performance; however, it is computationally expensive and inefficient. Parameter-efficient fine-tuning (PEFT) methods, like LoRA, have been proposed to address these challenges by freezing the pre-trained model and adding lightweight task-specific modules. LoRA, in particular, has proven effective, but its application to multi-task scenarios is limited by interference between tasks. Recent approaches, such as Mixture-of-Experts (MOE) and asymmetric LoRA, have aimed to mitigate these issues but still struggle with sample scarcity and noise interference due to   their fixed structure. In response, we propose CoLA, a more flexible LoRA architecture with an efficient initialization scheme, and introduces three collaborative strategies to enhance performance by better utilizing the quantitative relationships between matrices $A$ and $B$. Our experiments demonstrate the effectiveness and robustness of CoLA, outperforming existing PEFT methods, especially in low-sample scenarios. Our data and code are fully publicly available\footnote{\url{https://github.com/zyy-2001/CoLA}}.

\end{abstract}

\section{Introduction}
\label{sec: introduction}

The scaling law~\cite{kaplan2020scaling, zhai2022scaling} of Large Language Models (LLMs) describes a power-law relationship between the performance of a deep learning model and its scale (\textit{e.g.}, number of \textbf{parameters}, \textbf{computation}, and \textbf{data}) as the model size increases. As the model scale grows, the rate of performance improvement gradually flattens. Despite the impressive understanding and expressive capabilities of LLMs, training such a model from scratch is costly, which hinders the application of the scaling law, especially for small companies or institutions. Fine-tuning a single LLM for different downstream tasks or knowledge domains has become a common paradigm in various vertical fields~\cite{araci2019finbert, peng2021mathbert, chalkidis2020legal, rasmy2021med}, and previous studies~\cite{lester2021power, hernandez2021scaling} indicate that the scaling law also applies to fine-tuning. However, this approach, \textit{i.e.}, full fine-tuning (FFT), requires entire pre-trained weights of the LLM to be involved in heavy gradient computation, demanding substantial computational resources and energy consumption, thus hindering further exploration of the scaling law. In response, parameter-efficient fine-tuning (PEFT) methods have been proposed, where the backbone model's parameters are frozen, and only a small number of additional parameters or external modules customized for specific tasks or multi-task learning are modified. Common methods include LoRA~\cite{hu2021lora}, Adapters~\cite{LesterAC21, P-tuning, IA3}, and other variants~\cite{liu2024dora, meng2024pissa, tian2024hydralora}, which offer a solution for companies and researchers with limited computational resources.

\begin{figure*}[t]
    \includegraphics[width=\linewidth]{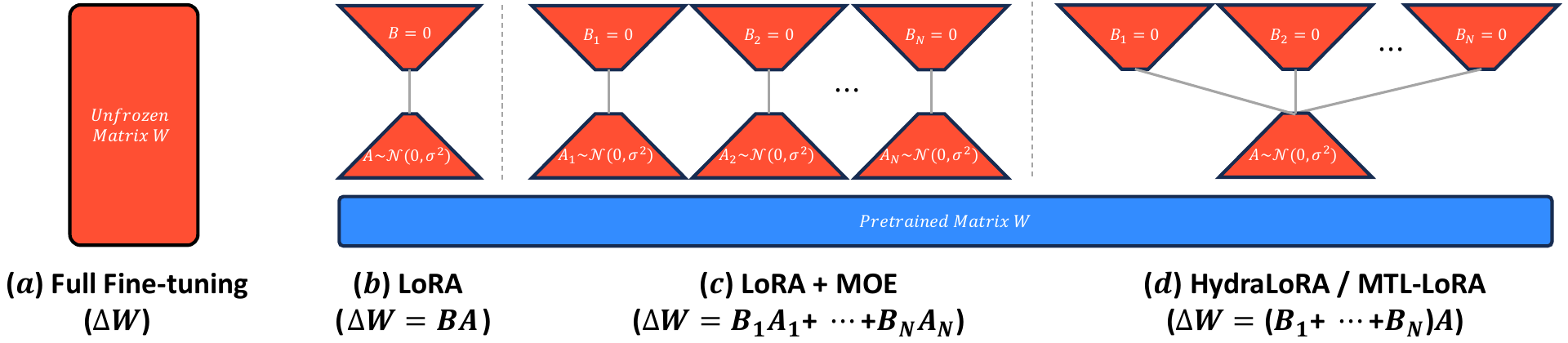}
    \vspace{-0.6cm}
    \textbf{\caption{The comparison between Full Fine-tuning and different LoRA variant structures.}
    \label{fig:related_work}}
    \vspace{-0.1cm}
\end{figure*}

As a PEFT method, LoRA has gained significant attention due to its simplicity, effectiveness, and generality, leading to many promising works, including exploration of better LoRA architectures. As shown in Figure~\ref{fig:related_work} (a) and (b), unlike full fine-tuning, which completely unfreezes the pre-trained weight matrix $W$, LoRA freezes the pre-trained matrix $W$ and approximates the incremental update $\Delta W$ of the pre-trained weights with two trainable low-rank matrices $A$ and $B$. These matrices are inserted into each layer of the pre-trained model, achieving comparable or even superior performance to full fine-tuning. However, a single LoRA module projects the features of different tasks into the same dense low-dimensional space, causing interference between tasks and failing to effectively separate the knowledge of different tasks, limiting adaptability in multi-task scenarios. Recent research~\cite{feng2024mixture, liu2024moe, agiza2024mtlora} has introduced the Mixture-of-Experts (MOE) idea to decouple multi-task information in LoRA, as shown in Figure~\ref{fig:related_work} (c). In this design, multiple experts are separately designed to learn task-shared and task-specific knowledge, with each expert consisting of a pair of low-rank matrices. This design allows the knowledge of multiple tasks to be effectively learned while maintaining parameter efficiency. Meanwhile, some studies~\cite{tian2024hydralora, yang2024mtl} have found that multiple smaller LoRA heads are more effective than a single LoRA, as matrix $A$ tends to learn the commonality across all data, while matrix $B$ focuses on the unique aspects of each intrinsic component. This leads to the design of an asymmetric LoRA architecture, as shown in Figure~\ref{fig:related_work} (d). However, a single matrix $A$ may struggle to capture commonality in limited samples and is prone to interference from noisy data. Moreover, these LoRA variant structures consistently initialize matrices $A$ and $B$ with Gaussian noise and zeros, which may result in small or random gradients early in training, slowing the fine-tuning process or causing the model to get stuck in suboptimal local minimum points. As a result, various alternative initialization schemes for LoRA have been explored~\cite{hayou2024impact, wang2024lora, meng2024pissa}, \textit{e.g.}, PiSSA~\cite{meng2024pissa}, which directly updates the model's principal components during fine-tuning through singular value decomposition, thereby accelerating convergence and improving performance.

Despite significant progress in the model parameter efficiency of different LoRA variant architectures, the scaling law for model fine-tuning remains limited in real-world scenarios due to the fact that sample labeling is expensive and private data is scarce. In particular, current LoRA methods, due to their fixed structure, fail to effectively capture the more complex inherent diversity in scarce samples. For example, as shown in Figure~\ref{fig:init} in Sec.~\ref{sec: results}, LoRA’s generalization ability deteriorates sharply when the sample size starts to decrease below 300. Therefore, it is crucial to further explore the numerical and collaborative relationships between matrices $A$ and $B$ in LoRA, freeing it from a monotonous structural design.

To address this, we introduce CoLA, a more flexible LoRA architecture, and extend the efficient PiSSA initialization scheme to CoLA, as shown in Figure~\ref{fig:cola} in Sec.\ref{sec: method}. CoLA does not enforce strict numerical relationships between matrices $A$ and $B$. To fully leverage their collaborative potential, we have designed three collaborative strategies with different energy consumption (\textbf{computation}): (i) Fully collaborative CoLA$^{\intercal}$: Different matrices $A$ and $B$ interact and learn from each other, with deep parameter sharing. (ii) Random collaborative CoLA$^{\dagger}$: A single matrix is randomly selected, without relying on specific combinations, allowing more diverse parameter learning. (iii) Heuristic collaborative CoLA$^{\ddagger}$: A combination of the two structures, integrating the different advantages of LoRA structures. The details of these three collaborative strategies are described in Sec.\ref{sec: strategy}. We have conducted extensive experiments to validate the effectiveness and robustness of CoLA, with experiments performed on two recent Llama models with different \textbf{parameter} sizes, using fine-tuning \textbf{data} that reflect realistic scenarios. Four interesting observations are made, as presented in Sec.~\ref{sec: results}.

\section{Related Works}
\label{sec: related_works}

\subsection{LoRA Architecture}

As a parameter-efficient fine-tuning (PEFT) method, LoRA has been widely used due to its simplicity, effectiveness, and generality. It effectively reduces the complexity of parameter updates by introducing low-rank decomposition, significantly improving fine-tuning efficiency while maintaining model performance. Previous research~\cite{qin2021exploring} has shown that despite the large number of parameters in pre-trained models, the intrinsic dimensionality of the model on downstream tasks is not large. Therefore, LoRA proposes the hypothesis of low-rank decomposition for the incremental update of pre-trained weight matrix $W_0\in \mathbb{R}^{n\times m}$:
\begin{flalign}
&W=W_0+\Delta W=W_0 + BA, \label{eq:lora}
\end{flalign}
where $B\in \mathbb{R}^{n\times r}, A\in \mathbb{R}^{r\times m}$, and $r\ll \min(n, m)$. It's noted that during the prediction phase, $W_0$, $A$ and $B$ can be combined into a single matrix without increasing the inference cost.

Due to its simplicity and practicality, the vanilla LoRA method has inspired considerable research. However, its simple structure sometimes struggles to effectively capture the diversity of data samples. As a result, some studies have attempted to combine the Mixture of Experts (MOE) approach, where different LoRA experts learn specific knowledge or tasks~\cite{liu2024moe, feng2024mixture, agiza2024mtlora}. In contrast, some methods are designed with the idea that the matrices $A$ and $B$ in LoRA tend to learn the commonality and intrinsic diversity of domain knowledge or tasks, leading to the proposal of asymmetric LoRA structures~\cite{yang2024mtl, tian2024hydralora}. The idea of using two matrices to learn commonalities and diversities is consistent with the layer-wise abstraction mechanism~\cite{lecun1998gradient, riesenhuber1999hierarchical, hinton2006fast, zhou2025revisiting, zhou2025disentangled, li2025mergenet} in deep learning. Inspired by this, a study~\cite{gao2024higher} combining MOE and LoRA has found that higher layers should allocate more experts to effectively learn the more complex features. However, in these LoRA methods, the matrices $A$ and $B$ are limited by a fixed numerical relationship, and the collaborative relationship between the matrices has not been explored further, leading to suboptimal performance.

\subsection{LoRA Parameter Initialization}
Typically, LoRA initializes the matrices $A$ and $B$ with Gaussian noise and zeros, respectively, to enforce $\Delta W=0$ at the beginning. Intuitively, initializing either the matrix $A$ or the matrix $B$ with zeros seems feasible, and empirical results~\cite{zhu2024asymmetry} indicate that both approaches achieve similar performance. However, other research~\cite{hayou2024impact} suggests that initializing the matrix $B$ with zeros generally leads to better results. Meanwhile, other fixed initializations instead of random initialization have been explored~\cite{meng2024pissa, ke2025unveiling}. Specifically, PiSSA~\cite{meng2024pissa} shows significant performance improvements on several tasks by initializing with top singular vectors to accelerate the convergence of LoRA, and many similar works have been proposed afterward~\cite{wang2024lora, wang2024lorapro, yang2024corda, lingam2024svft, zhang2024spectral}. Based on the effectiveness of singular value decomposition, we extend PiSSA to the proposed CoLA, and it serves as an essential component, as discussed in Observation 2 of Sec.~\ref{sec: results}.

\section{CoLA}
\label{sec: method}

\begin{figure*}[t]
    \includegraphics[width=\linewidth]{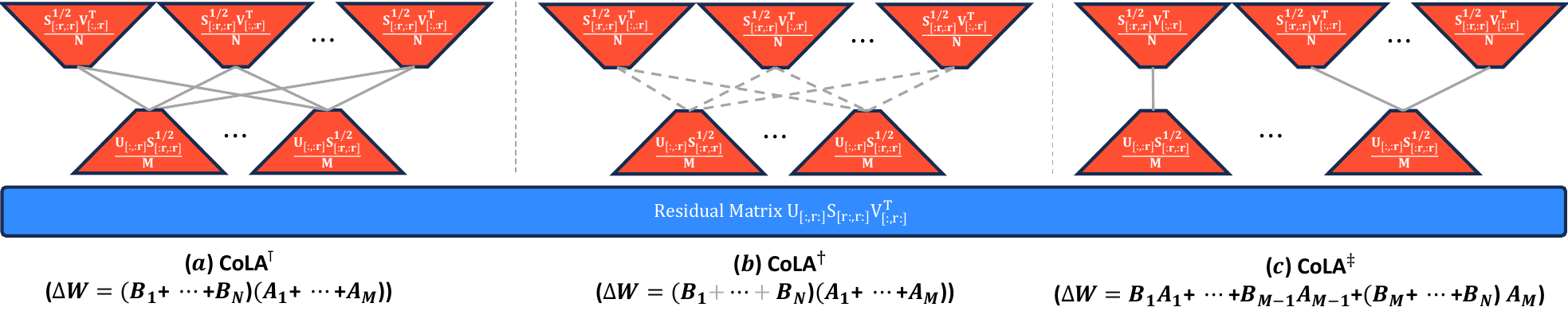}
    \vspace{-0.6cm}
    \textbf{\caption{Overview of CoLA with three collaborative strategies.}
    \label{fig:cola}}
\end{figure*}

In this section, we introduce the proposed CoLA, a flexible LoRA achitecture as illustrated in Figure~\ref{fig:cola}. We then present the extended PiSSA initialization scheme applied to CoLA and provide three collaborative strategies for the matrices $A$ and $B$.

\subsection{Flexible LoRA Architecture}

Previous LoRA architectures have typically been constrained by fixed relationships between the number of matrices $A$ and $B$. Specifically, as shown in Figures~\ref{fig:related_work} (b) and (c), in vanilla LoRA and traditional LoRA + MOE architectures~\cite{liu2024moe}, the setting is $\text{\#}A = \text{\#}B = N$, where $\text{\#}A$ and $\text{\#}B$ denote the number of matrices $A$ and $B$, and $N$ is a hyperparameter representing the number of experts. However, this symmetric structure may struggle to effectively learn both the shared components and intrinsic diversity of domain knowledge. To address this, some recent works have proposed asymmetric LoRA architectures. As illustrated in Figure~\ref{fig:related_work} (d), HydraLoRA~\cite{tian2024hydralora} and MTL-LoRA~\cite{yang2024mtl} adopt the setting $\text{\#}A = 1, \text{\#}B = N$. However, a single matrix $A$ may fail to capture commonalities in limited samples and is prone to data noise, especially in data-scarce real-world scenarios. In response, we introduce a more flexible LoRA architecture—CoLA—that frees itself from the fixed relationship between the number of matrices, \textit{i.e.}, $\text{\#}A = M, \text{\#}B = N$, where $M$ is also a hyperparameter, as illustrated in Figure~\ref{fig:cola}. Notably, existing LoRA architectures can be viewed as special cases of CoLA.

\subsection{Extended PiSSA}

For any matrix $W\in \mathbb{R}^{n\times m}$, a singular value decomposition (SVD) of the following form can be found:
\begin{flalign}
&W=USV^\top, \label{eq:svd}
\end{flalign}
where $U\in \mathbb{R}^{m\times m}$ and $V\in \mathbb{R}^{n\times n}$ are orthogonal matrices, and $V^\top$ is the transpose of $V$. $S\in \mathbb{R}^{n\times m}$ is a non-negative diagonal matrix:
\begin{flalign}
S_{i, j}= \begin{cases}\Lambda_i, & i=j \\ 0, & i \neq j\end{cases}
\end{flalign}
where the diagonal elements are in descending order, \textit{i.e.}, $\Lambda_1\geq \Lambda_2 \geq\cdots \geq 0$, known as the singular values.

PiSSA divides $S$, along with $U$ and $V$, into two groups: the principal singular values and vectors ($\left\{U_{[:, :r]}, S_{[:r, :r]}, V_{[;, :r]}\right\}$) and the residual singular values and vectors ($\left\{U_{[:, r:]}, S_{[r:, r:]}, V_{[;, r:]}\right\}$). The principal singular values and vectors are used to initialize the matrices $A$ and $B$ in LoRA:
\begin{flalign}
& A=U_{[:, :r]} S_{[:r, :r]}^{1/2}, \quad B=S_{[:r, :r]}^{1/2} V_{[:, :r]}^\top.
\end{flalign}

Meanwhile, the residual singular values and vectors are used to construct the residual matrix, which remains frozen during fine-tuning:
\begin{flalign}
& W_0 = U_{[:, r:]} S_{[r:, r:]} V_{[:, r:]}^\top.
\end{flalign}

Since elements of $\Lambda_{[:r]}$ $\gg$ elements of $\Lambda_{[r:]}$, PiSSA assumes that the initial $BA$ contains the most important directions of $W$, leading to faster and better convergence. This assumption is supported by the Eckart-Young-Mirsky theorem, which is ignored by PiSSA, as formally described in Theorem~\ref{thm:young}.

\begin{tcolorbox}[colback=gray!10,colframe=gray!50]
\begin{theorem}\label{thm:young}
If the SVD of $W \in \mathbb{R}^{n \times m}$ is $U S V^{\top}$, then the optimal rank $r$ approximation of $W$ is $U_{[:n, :r]} S_{[: r, :r]} V_{[:m, :r]}^{\top}$.
\end{theorem}
\end{tcolorbox}

We extend PiSSA to the flexible CoLA architecture. For the principal singular values and vectors of matrices $A$ and $B$, $U_{[:, :r]} S_{[:r, :r]}^{1/2}$ and $S_{[:r, :r]}^{1/2} V_{[:, :r]}^T$, which are evenly distributed to each matrix $A_i$ ($1 \leq i \leq M$) and $B_j$ ($1 \leq j \leq N$):
\begin{flalign}
& A_i=\frac{U_{[:, :r]} S_{[:r, :r]}^{1/2}}{M}, \quad B_j=\frac{S_{[:r, :r]}^{1/2} V_{[:, :r]}^T}{N}.
\end{flalign}

Each matrix $A_i$ and $B_j$ is initially treated equally and aligned for full fine-tuning. During the fine-tuning process, each matrix is optimized in different directions, which allows CoLA to have more diverse generalization capabilities. The extended PiSSA initialization significantly benefits CoLA, as observed in Observation 2 of Sec.~\ref{sec: results}.

\begin{table*}[t]
\resizebox{\textwidth}{!}{%
\begin{tabular}{@{}c|c|c|c@{}}
\toprule[1.5pt]
\textbf{Domain}        & \textbf{Fine-tuning Dataset}                                                                                                & \textbf{Benchmark}                                                               & \textbf{Function}                                                                   \\ \midrule
\textbf{Generality}    & databricks-dolly-15k~\cite{conover2023free}                                                  & MMLU~\cite{hendrycks2020measuring}                & General Instruction Following                                              \\
\textbf{Law}           & Lawyer-Instruct~\cite{Lawyer-Instruct} and US-Terms~\cite{legal_lama} & Legal Tasks in MMLU                                                     & Legal Judgment                                                             \\
\textbf{Medicine}      & GenMedGPT-5k and clinic-10k from ChatDoctor~\cite{chatdoctor}                                   & Medical Tasks in MMLU                                                   & Medical Diagnosis                                                          \\
\textbf{Math}          & Training Set of GSM8k~\cite{gsm8k}                                                           & Test Set of GSM8k                                                       & Mathematical Reasoning                                                     \\
\textbf{Finance}       & Training Set of fingpt-fineval~\cite{wang2023fingptbenchmark}                   & Test Set of fingpt-fineval                                              & Financial Q\&A                                                             \\ \midrule
\textbf{Multi-tasking} & OpenOrca~\cite{lian2023openorca}                                                             & Big-Bench Hard (BBH)~\cite{suzgun2022challenging} & Natural Language Understanding (NLU) and Natural Language Generation (NLG) \\ \bottomrule[1.5pt]
\end{tabular}%
}
\vspace{-0.2cm}
\caption{The basic information of the datasets used in our experiments.}
\label{tab: dataset}
\end{table*}

\subsection{Collaborative Strategy}
\label{sec: strategy}
In the vanilla LoRA method, the matrices $A$ and $B$ represent the incremental update $\Delta W = BA$ of pre-trained weights through a one-to-one relationship. However, this simple, singular connection fails to capture the diversity of different tasks. Therefore, some LoRA methods introduce the idea of MOE (Mixture of Experts) to construct multiple distinct one-to-one relationships within the matrices $A$ and $B$ to achieve the more diverse incremental update $\Delta W = B_1A_1 + \cdots + B_NA_N$. However, since each expert is independent, although it can effectively learn the intrinsic diversity of knowledge, it struggles to capture the commonality of domain-specific knowledge. Thus, HydraLoRA and MTL-LoRA methods adopt a one-to-many relationship between the matrices $A$ and $B$ to represent the more complex incremental update $\Delta W = (B_1 + \cdots + B_N)A$. However, the knowledge that a single $A$ matrix can learn is limited and more prone to noise, especially in real-world scenarios with sparse samples. Based on this, we introduce a many-to-many relationship within the matrices $A$ and $B$ to represent the more refined incremental update, as shown in Figure~\ref{fig:cola}. The three collaborative strategies are as follows:
\begin{itemize}
    \item \textbf{Fully Collaborative CoLA$^{\intercal}$}: CoLA$^{\intercal}$ represents the finest incremental update $\Delta W = (B_1 + \cdots + B_N)(A_1 + \cdots + A_M)$ by combining each matrix $A$ and $B$. This collaborative strategy breaks down the information transmission barrier between each matrix $A$ and $B$, making it easier for beneficial knowledge to be shared. However, this may introduce more energy consumption.
    \item \textbf{Random Collaborative CoLA$^{\dagger}$}: Inspired by the dropout regularization technique~\cite{srivastava2014dropout} in deep learning, CoLA$^{\dagger}$ represents the more robust incremental update $\Delta W = (B_1\textcolor{gray!45}{\textbf{+}} \cdots \textcolor{gray!45}{\textbf{+}} B_N)(A_1 + \cdots + A_M)$ by combining each matrix $A$ with a randomly chosen matrix $B$. This collaborative strategy does not rely on a specific combination, making the learned knowledge expected to be more robust, while incurring the fewest energy consumption.
    \item \textbf{Heuristic Collaborative CoLA$^{\ddagger}$}: CoLA$^{\ddagger}$ integrates multiple one-to-one and one-to-many relationships between matrices $A$ and $B$ (which can be seen as the combination of Figures~\ref{fig:related_work} (c) and (d)) to represent the complex and diverse incremental update $\Delta W = B_1A_1 + \cdots + B_{M-1}A_{M-1} + (B_{M} + \cdots + B_N)A_M$ (assume $M < N$). This collaborative strategy combines the advantages of two specific combinations, enabling the learning of both general and diverse knowledge, while producing moderate energy consumption.
\end{itemize}

We analyze the energy consumption produced by these three collaborative strategies in Observation 4 of Sec.~\ref{sec: results}.

\begin{table*}[t]
\resizebox{\textwidth}{!}{%
\begin{tabular}{@{}cl|c|c|c|c|cl|cl|cl|cl|cl|cl|cl|cl|cl|cl|cl@{}}
\toprule[1.5pt]
\multicolumn{2}{c|}{\textbf{Method}}                            & \textbf{Llama-3.1-8B} & \textbf{Prompt Tuning} & \textbf{P-Tuning} & \textbf{IA$^{3}$} & \multicolumn{2}{c|}{\textbf{LoRA$_{r=8}$}}         & \multicolumn{2}{c|}{\textbf{LoRA$_{r=16}$}}        & \multicolumn{2}{c|}{\textbf{LoRA$_{r=24}$}}        & \multicolumn{2}{c|}{\textbf{LoRA$_{r=32}$}}        & \multicolumn{2}{c|}{\textbf{DoRA}}                  & \multicolumn{2}{c|}{\textbf{PiSSA}}                 & \multicolumn{2}{c|}{\textbf{HydraLoRA}}             & \multicolumn{2}{c|}{\textbf{CoLA}}                            & \multicolumn{2}{c|}{\textbf{CoLA$^{\intercal}$}}                     & \multicolumn{2}{c|}{\textbf{CoLA$^{\dagger}$}}                           & \multicolumn{2}{c}{\textbf{CoLA$^{\ddagger}$}}                     \\ \midrule
\rowcolor[HTML]{EFEFEF} 
\multicolumn{2}{c|}{\cellcolor[HTML]{EFEFEF}\textbf{\#A \big| \#B}} & -                     & -                      & -                 & -                 & \multicolumn{2}{c|}{\cellcolor[HTML]{EFEFEF}1 \,\,\, \big| \,\,\, 1}      & \multicolumn{2}{c|}{\cellcolor[HTML]{EFEFEF}1\,\,\,\, \big| \,\,\,\, 1}      & \multicolumn{2}{c|}{\cellcolor[HTML]{EFEFEF}1\,\,\,\, \big| \,\,\,\, 1}      & \multicolumn{2}{c|}{\cellcolor[HTML]{EFEFEF}1\,\,\,\, \big| \,\,\,\, 1}      & \multicolumn{2}{c|}{\cellcolor[HTML]{EFEFEF}1\,\, \big| \,\,1}      & \multicolumn{2}{c|}{\cellcolor[HTML]{EFEFEF}1\,\, \big| \,\,1}      & \multicolumn{2}{c|}{\cellcolor[HTML]{EFEFEF}1 \,\,\,\, \big|\,\,\,\,\, 3}      & \multicolumn{2}{c|}{\cellcolor[HTML]{EFEFEF}1\,\, \big| \,\, 3}                & \multicolumn{2}{c|}{\cellcolor[HTML]{EFEFEF}2 \, \big|\,\, 
 3}          & \multicolumn{2}{c|}{\cellcolor[HTML]{EFEFEF}2\,\, \big|\,\, 
 3}                & \multicolumn{2}{c}{\cellcolor[HTML]{EFEFEF}2\,\, \big|\,\, 
 3}          \\ \midrule
\rowcolor[HTML]{EFEFEF} 
\multicolumn{2}{c|}{\cellcolor[HTML]{EFEFEF}\textbf{\%Param}}   & -                     & 0.0004                 & 0.0280            & 0.0065            & \multicolumn{2}{c|}{\cellcolor[HTML]{EFEFEF}0.2605} & \multicolumn{2}{c|}{\cellcolor[HTML]{EFEFEF}0.5196} & \multicolumn{2}{c|}{\cellcolor[HTML]{EFEFEF}0.7774} & \multicolumn{2}{c|}{\cellcolor[HTML]{EFEFEF}1.0338} & \multicolumn{2}{c|}{\cellcolor[HTML]{EFEFEF}0.2605} & \multicolumn{2}{c|}{\cellcolor[HTML]{EFEFEF}0.2605} & \multicolumn{2}{c|}{\cellcolor[HTML]{EFEFEF}0.5785} & \multicolumn{2}{c|}{\cellcolor[HTML]{EFEFEF}0.5325}           & \multicolumn{2}{c|}{\cellcolor[HTML]{EFEFEF}0.6551}     & \multicolumn{2}{c|}{\cellcolor[HTML]{EFEFEF}0.6551}           & \multicolumn{2}{c}{\cellcolor[HTML]{EFEFEF}0.6551}     \\ \midrule
\multicolumn{2}{c|}{\textbf{Generality}}                        & 22.95                 & 25.38                  & 24.68             & 22.96             & \multicolumn{2}{c|}{50.36}                          & \multicolumn{2}{c|}{51.47}                          & \multicolumn{2}{c|}{52.91}                          & \multicolumn{2}{c|}{52.35}                          & \multicolumn{2}{c|}{51.56}                          & \multicolumn{2}{c|}{\underline{54.72}}                          & \multicolumn{2}{c|}{45.86}                          & \multicolumn{2}{c|}{\cellcolor[HTML]{CBCEFB}\textbf{58.04**}} & \multicolumn{2}{c|}{\cellcolor[HTML]{CBCEFB}\textbf{58.21**}} & \multicolumn{2}{c|}{\cellcolor[HTML]{CBCEFB}\textbf{42.26}} & \multicolumn{2}{c}{\cellcolor[HTML]{CBCEFB}\textbf{57.08**}} \\
\multicolumn{2}{c|}{\textbf{Law}}                               & 24.62                 & 25.75                  & 26.09             & 24.62             & \multicolumn{2}{c|}{25.98}                          & \multicolumn{2}{c|}{\underline{26.60}}                          & \multicolumn{2}{c|}{24.96}                          & \multicolumn{2}{c|}{25.92}                          & \multicolumn{2}{c|}{26.32}                          & \multicolumn{2}{c|}{26.58}                          & \multicolumn{2}{c|}{26.26}                          & \multicolumn{2}{c|}{\cellcolor[HTML]{CBCEFB}\textbf{36.25**}} & \multicolumn{2}{c|}{\cellcolor[HTML]{CBCEFB}\textbf{41.46**}} & \multicolumn{2}{c|}{\cellcolor[HTML]{CBCEFB}\textbf{26.27}} & \multicolumn{2}{c}{\cellcolor[HTML]{CBCEFB}\textbf{31.04*}} \\
\multicolumn{2}{c|}{\textbf{Medicine}}                          & 23.82                 & 24.91                  & 25.12             & 23.82             & \multicolumn{2}{c|}{42.66}                          & \multicolumn{2}{c|}{47.17}                          & \multicolumn{2}{c|}{\underline{50.38}}                          & \multicolumn{2}{c|}{45.94}                          & \multicolumn{2}{c|}{43.28}                          & \multicolumn{2}{c|}{44.64}                          & \multicolumn{2}{c|}{40.61}                          & \multicolumn{2}{c|}{\cellcolor[HTML]{CBCEFB}\textbf{56.11**}} & \multicolumn{2}{c|}{\cellcolor[HTML]{CBCEFB}\textbf{54.33**}} & \multicolumn{2}{c|}{\cellcolor[HTML]{CBCEFB}\textbf{41.24}} & \multicolumn{2}{c}{\cellcolor[HTML]{CBCEFB}\textbf{50.23}} \\
\multicolumn{2}{c|}{\textbf{Math}}                              & 24.79                 & 26.00                  & 26.16             & 24.72             & \multicolumn{2}{c|}{51.02}                          & \multicolumn{2}{c|}{54.66}                          & \multicolumn{2}{c|}{56.94}                          & \multicolumn{2}{c|}{56.48}                          & \multicolumn{2}{c|}{51.18}                          & \multicolumn{2}{c|}{\underline{57.00}}                          & \multicolumn{2}{c|}{47.31}                          & \multicolumn{2}{c|}{\cellcolor[HTML]{CBCEFB}\textbf{57.71}}  & \multicolumn{2}{c|}{\cellcolor[HTML]{CBCEFB}\textbf{59.14**}} & \multicolumn{2}{c|}{\cellcolor[HTML]{CBCEFB}\textbf{45.96}} & \multicolumn{2}{c}{\cellcolor[HTML]{CBCEFB}\textbf{56.34}} \\
\multicolumn{2}{c|}{\textbf{Finance}}                           & 26.42                 & 28.30                  & 21.51             & 26.42             & \multicolumn{2}{c|}{40.38}                          & \multicolumn{2}{c|}{44.53}                          & \multicolumn{2}{c|}{45.66}                          & \multicolumn{2}{c|}{\underline{48.68}}                          & \multicolumn{2}{c|}{41.13}                          & \multicolumn{2}{c|}{46.79}                          & \multicolumn{2}{c|}{38.87}                          & \multicolumn{2}{c|}{\cellcolor[HTML]{CBCEFB}\textbf{52.45**}} & \multicolumn{2}{c|}{\cellcolor[HTML]{CBCEFB}\textbf{50.19**}} & \multicolumn{2}{c|}{\cellcolor[HTML]{CBCEFB}\textbf{37.51}} & \multicolumn{2}{c}{\cellcolor[HTML]{CBCEFB}\textbf{45.32}} \\ \bottomrule[1.5pt]
\end{tabular}%
}
\vspace{-0.2cm}
\caption{Comparison of 0-shot performance (\%) of different fine-tuning methods based on Llama-3.1-8B across multiple single domains. The experiments are repeated 5 times under random seeds 42 to 46 and the average performance is reported. $\text{\#}A$ and $\text{\#}B$ represent the number of matrices $A$ and $B$, respectively. * and ** indicate that the improvements over \underline{the strongest baseline} with underlined are statistically significant, with p \textless 0.05 and p \textless 0.01, respectively. The results based on Llama-3.2-3B are in Appendix~\ref{apx: results}.}
\label{tab:single_domain}
\end{table*}

\begin{table*}[t]
\centering
\resizebox{0.78\textwidth}{!}{%
\begin{tabular}{@{}cc|c|c|c|c|c|c|c|c|c|c@{}}
\toprule[1.5pt]
\multicolumn{2}{c|}{\textbf{Method}}                                        & \textbf{Base} & \textbf{LoRA$_{r=64}$} & \textbf{MOELoRA}               & \textbf{MTL-LoRA}              & \textbf{MoLA} & \textbf{HydraLoRA} & \textbf{CoLA}                          & \textbf{CoLA$^{\intercal}$}                         & \textbf{CoLA$^{\dagger}$}                     & \textbf{CoLA$^{\ddagger}$}                     \\ \midrule
\rowcolor[HTML]{EFEFEF} 
\multicolumn{2}{c|}{\cellcolor[HTML]{EFEFEF}\textbf{\#A \big| \#B}}             & -             & 1\,\,\,\, \big| \,\,\,\,1                   & 8\,\,\,\,\, \big| \,\,\,\,\,8                          & 1\,\,\,\,\, \big| \,\,\,\,\,14                         & $\tilde{8}$\,\, \big| \,\,$\tilde{8}$         & 1\,\,\,\,\, \big| \,\,\,\,\,14             & \cellcolor[HTML]{EFEFEF}1\,\, \big| \,\,14         & 4\,\, \big| \,\,10                                 & \cellcolor[HTML]{EFEFEF}4\,\, \big| \,\,10     & \multicolumn{1}{c}{\cellcolor[HTML]{EFEFEF}4\,\, \big| \,\,10}     \\ \midrule
\rowcolor[HTML]{EFEFEF} 
\multicolumn{2}{c|}{\cellcolor[HTML]{EFEFEF}\textbf{\%Param}}               & -             & 2.0465                  & \cellcolor[HTML]{EFEFEF}2.0482 & \cellcolor[HTML]{EFEFEF}2.0051 & 2.1654        & 2.2107             & 2.0026                                 & 1.8330                                 & 1.8330                             & \multicolumn{1}{c}{1.8330}                             \\ \midrule
\multicolumn{1}{c|}{}                                         & \textbf{Llama-3.2-3B} & 29.36         & 34.89                   & 30.77                          & 32.31                          & \underline{35.11}         & 29.64              & \cellcolor[HTML]{CBCEFB}\textbf{36.87**} & \cellcolor[HTML]{CBCEFB}\textbf{36.47*} & \cellcolor[HTML]{CBCEFB}\textbf{31.18} & \multicolumn{1}{c}{\cellcolor[HTML]{CBCEFB}\textbf{34.58}} \\
\multicolumn{1}{c|}{\multirow{-2}{*}{\textbf{Multi-tasking}}} & \textbf{Llama-3.1-8B} & 29.47         & \underline{42.99}                   & 40.53                          & 41.39                          & 41.16         & 39.08              & \cellcolor[HTML]{CBCEFB}\textbf{42.87} & \cellcolor[HTML]{CBCEFB}\textbf{43.62*} & \cellcolor[HTML]{CBCEFB}\textbf{39.26} & \multicolumn{1}{c}{\cellcolor[HTML]{CBCEFB}\textbf{42.16}} \\ \bottomrule[1.5pt]
\end{tabular}%
}
\vspace{-0.2cm}
\caption{Comparison of 0-shot performance (\%) of different fine-tuning methods across multiple domains.}
\label{tab:multiple_domain}
\end{table*}

\section{Experiments}
\label{sec: experiment}
In this section, we present the setup and details of the experiments. Then, we share our findings and provide concise explanations.

\subsection{Experimental Setup}

\subsubsection{Datasets and Benchmarks}
Following HydraLoRA~\cite{tian2024hydralora}, we evaluate the performance of different fine-tuning methods on datasets from single and multiple domains. Table~\ref{tab: dataset} shows the basic information of these datasets, and more details can be found in Appendix~\ref{apx: datasets}.

\begin{figure*}[t]
    \includegraphics[width=\linewidth]{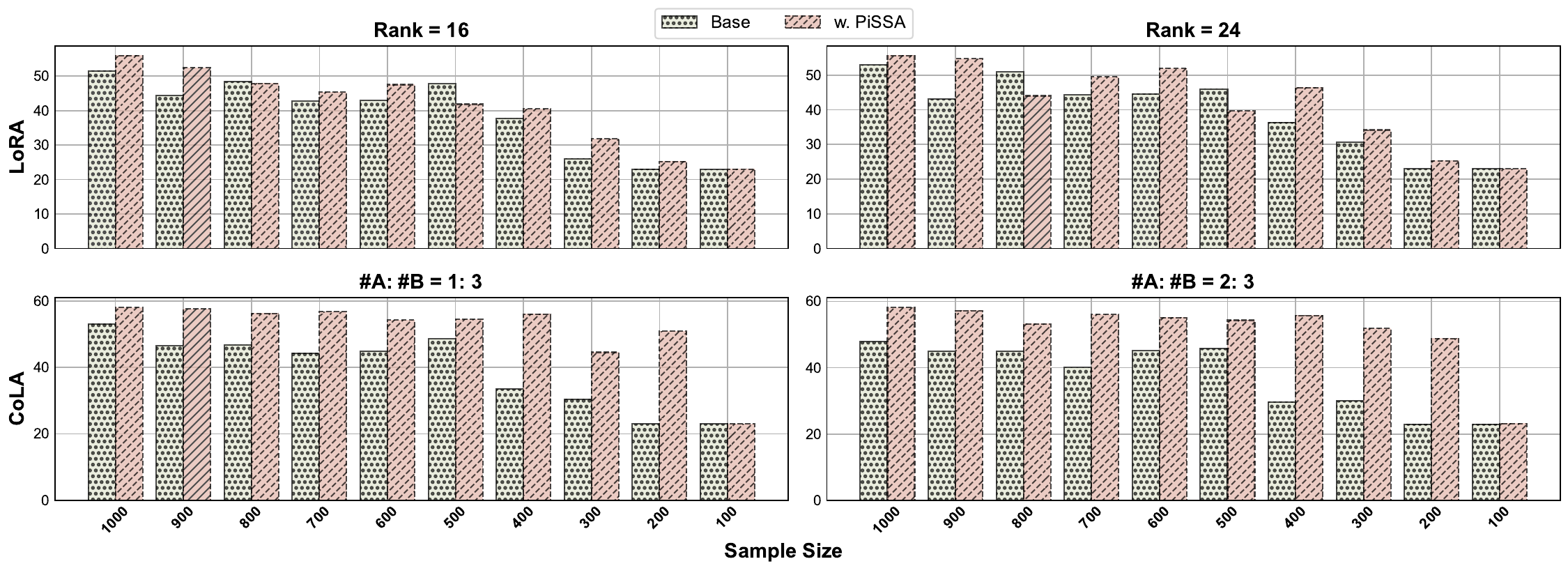}
    \vspace{-0.7cm}
    \textbf{\caption{The impact of PiSSA initialization on LoRA and CoLA based on Llama-3.1-8B in the generality domain when the sample size is reduced.}
    \label{fig:init}}
\end{figure*}

\subsubsection{Baselines}

We select recent Llama models with different parameter scales~\cite{dubey2024llama} (Llama-3.2-3B and Llama-3.1-8B) as the backbone models for optimization. Additionally, we compare CoLA with several different PEFT methods:
\begin{itemize}
\item {Single domain: Full fine-tuning (FFT) (not applied to Llama-3.1-8B due to resource limitations), Prompt Tuning~\cite{LesterAC21}, P-Tuning~\cite{P-tuning}, IA$^{3}$~\cite{IA3}, LoRA~\cite{hu2021lora}, DoRA~\cite{liu2024dora}, PiSSA~\cite{meng2024pissa}, HydraLoRA~\cite{tian2024hydralora}.}
\item {Multiple domain: MOELoRA~\cite{liu2024moe}, MTL-LoRA~\cite{yang2024mtl}, MoLA~\cite{gao2024higher}, HydraLoRA.}
\end{itemize}

Details of the aforementioned PEFT methods can be found in Appendix~\ref{apx: baselines}.  Specifically, unless stated otherwise, the LoRA rank is set to 8 by default for the PEFT methods related to LoRA.

\begin{figure*}[!ht]
    \includegraphics[width=\linewidth]{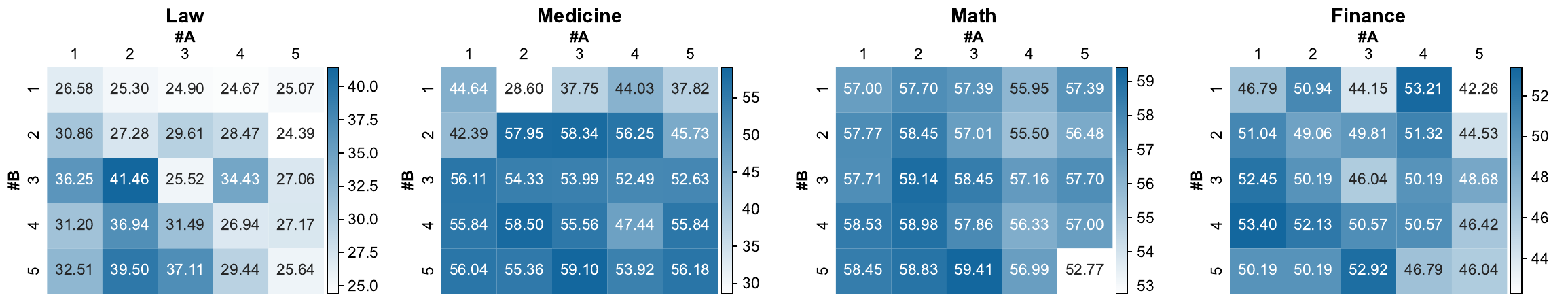}
    \vspace{-0.6cm}
    \textbf{\caption{The performance of Llama-3.1-8B when the number of matrices $A$ and $B$ in CoLA differs across the domains of law, medicine, math, and finance.}
    \label{fig:num}}
    \vspace{-0.3cm}
\end{figure*}

\subsubsection{Implementation}

Previous studies have shown that the quality of text generated by models can be significantly influenced by sampling strategies (\textit{e.g.}, temperature), and that generating the entire text leads to inefficiencies~\cite{renze2024effect, zhu2024hot, patel2024exploring}. Meanwhile, the multiple-choice inference mode generates stable and consistent results by using the model's logits, requiring only the computation of the logits for the final token~\cite{hendryckstest2021}. The specific differences between these two inference modes, as well as the code, can be found in Appendix~\ref{apx: modes}. To ensure fairness and reproducibility in evaluation, we follow the evaluation settings recommended in LlamaFactory~\cite{zheng2024llamafactory} and uniformly convert the model's generation task into a classification task. However, this may conflict with the original instruction setup. We use powerful language models to normalize the conflicting datasets (GSM8K and BBH) into multiple-choice formats. All experiments are conducted using the LlamaFactory framework. The prompts for normalization, the prompt template used for the model's zero-shot evaluation, and additional experimental details can be found in Appendix~\ref{apx: details}. 

\subsection{Results}
\label{sec: results}
We conduct extensive experiments to demonstrate the value of the proposed CoLA. Four key observations are summarized as follows.

\begin{githubquote}
    \textbf{Observation 1.} CoLA is effective on both single and multiple knowledge domains.
\end{githubquote}

Table~\ref{tab:single_domain} and Table~\ref{tab:single_domain_other} in Appendix~\ref{apx: results} show the performance comparison of different fine-tuning methods based on Llama-3.1-8B and Llama-3.2-3B across multiple single domains in a 0-shot setting. Our findings are as follows: (1) Compared to methods like LoRA, fine-tuning approaches including Prompt Tuning, P-Tuning, and IA$^{3}$, although designed with fewer parameters, struggle to effectively capture the patterns of few samples and generalize to more samples in scenarios where data is scarce, which can become a curse in practical settings. (2) LoRA remains a simple yet hard-to-beat baseline compared to the base model and other methods, consistently achieving top-2 performance across these baselines. (3) DoRA generally improves the performance of LoRA$_{r=8}$, indicating the effectiveness of updating the weight decomposition's directional components in LoRA. (4) PiSSA achieves impressive performance with fewer parameters, due to its ability to pre-learn the principal components of the pre-trained weights for faster convergence. This also confirms the correctness of CoLA's efficient initialization scheme. Meanwhile, HydraLoRA does not achieve the expected results, likely due to its Gaussian noise initialization, which may lead to overfitting under conditions with scarce samples. (5) CoLA, CoLA$^{\intercal}$, and CoLA$^{\ddagger}$ (especially CoLA and CoLA$^{\intercal}$) consistently achieve excellent performance, which is closely tied to the advantages analyzed in Sec.~\ref{sec: strategy}, while CoLA$^{\dagger}$ does not yield the expected results. This pseudo-LoRA implementation, averaging a matrix $B$, contradicts the quantitative relationship between the matrices $A$ and $B$ identified in Observation 3.

Table~\ref{tab:multiple_domain} presents a performance comparison of different fine-tuning methods across multiple domains in a 0-shot setting. Several insights are drawn: (1) Compared to the base model, the three multi-task LoRA method (MOELoRA, MTL-LoRA, and HydraLoRA) are effective, but fail to effectively learn the instruction patterns in the pre-trained weights due to the random initialization of their experts. (2) MoLA (specifically MoLA-$\nabla$) and LoRA$_{r=64}$ show comparable performance and consistently outperform MOELoRA, indicating the effectiveness of allocating more experts to higher layers with advanced features. (3) With the similar parameters, CoLA and CoLA$^{\intercal}$ consistently achieve the best performance, which indicates that they can also learn the inherent diversity of domain-specific knowledge in multitask learning. This is due to the fully collaborative strategy, which allows each matrix $A$ and $B$ to fully share knowledge.

\begin{figure*}[t]
    \includegraphics[width=\linewidth]{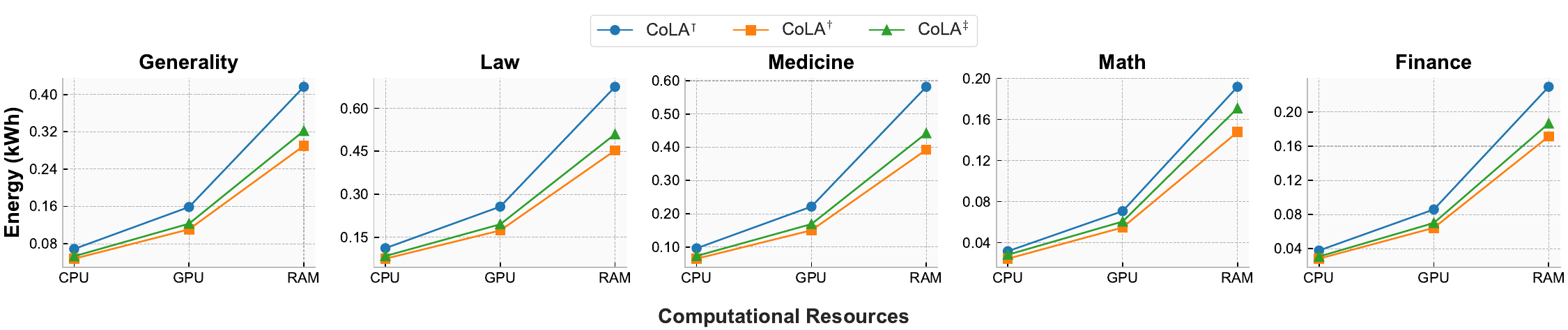}
    \vspace{-0.6cm}
    \textbf{\caption{Energy consumption of three collaborative strategies based on Llama-3.1-8B.}
    \label{fig:energy}}
    \vspace{-0.3cm}
\end{figure*}

\begin{githubquote}
    \textbf{Observation 2.} Compared to the LoRA method, the impact of parameter initialization on CoLA is more significant when the sample size is reduced.
\end{githubquote}
\label{ob: 2}
We investigate the robustness of the CoLA initialization scheme with continuously decreasing samples in harsh environments. In the generality domain, based on Llama-3.1-8B, we select LoRA$_{r=16}$ and LoRA$_{r=24}$ as the baselines for our CoLA ($\text{\#}A=1, \text{\#}B=3$) and CoLA$^{\intercal}$ ($\text{\#}A=2, \text{\#}B=3$), respectively, to explore the necessity of the extended PiSSA initialization for CoLA, as shown in Figure~\ref{fig:init}. From this, we observe: (1) In extremely harsh environments (sample size = 100), all methods fail, especially the LoRA, whose performance starts to degrade sharply when the sample size reaches 300. (2) CoLA without the extended PiSSA, while not outperforming LoRA, demonstrates consistently the best performance after initialization and remains stable even with fewer samples (sample size = 200). This suggests that the extended PiSSA has a significant impact on CoLA. CoLA's ability to maintain high performance in such challenging environments is attributed to its use of multiple different matrices $A$ and $B$, and the extended PiSSA initialization allows both matrices to learn the foundational instruction patterns of the pre-trained model, enabling efficient learning in distinct directions, leading to improved generalization.

\begin{githubquote}
    \textbf{Observation 3.} In CoLA, the number of matrix $A$ should be fewer than the number of matrix $B$, as the benefit of increasing the matrix $B$ outweighs that of increasing the matrix $A$.
\end{githubquote}

We explore the quantitative relationship between matrices $A$ and $B$ in CoLA across four domains: law, medicine, math, and finance. The experiments are based on Llama-3.1-8B, where $\text{\#}A$ and $\text{\#}B$ range from 1 to 5. From Figure~\ref{fig:num}, we can observe:
\begin{itemize}
    \item When either matrix $A$ or $B$ is fixed, increasing the number of the other matrix generally benefits the model. However, excessive increase can lead to overfitting (\textit{e.g.}, $\text{\#}A=5, \text{\#}B=1$ in the finance domain), highlighting the need for careful tuning of the number of $A$ and $B$ matrices.
    \item When $\text{\#}A=\text{\#}B$, simply increasing the number of experts does not necessarily improve model performance and may even degrade it (\textit{e.g.}, $\text{\#}A=\text{\#}B=5$ in the math domain). This suggests that traditional LoRA + MOE architectures, like MOELoRA, may not be optimal. On the contrary, when matrices $A$ and $B$ are asymmetric, particularly when $\text{\#}A<\text{\#}B$, increasing the number of experts tends to benefit the model (\textit{e.g.}, $\text{\#}A=1, \text{\#}B=3 \rightarrow \text{\#}A=2, \text{\#}B=4 \rightarrow \text{\#}A=3, \text{\#}B=5$).
    \item Let $x < y$. The model benefits more from $\text{\#}A=x, \text{\#}B=y$ rather than $\text{\#}A=y, \text{\#}B=x$. That is, the number of matrix $A$ in CoLA should be fewer than that of matrix $B$. This insight is inspired by the model structure designs in HydraLoRA and MoLA-$\nabla$: matrix $A$ learns the underlying commonalities in the data, while matrix $B$ focuses on the unique aspects of each component, with higher-level features receiving more weight. This can also be drawn from real life, where people often remember the contour of a face but overlook facial details (\textit{e.g.}, the width of the nose), though these details are crucial for accurate facial recognition~\cite{zhao2003face, liu2010fusion, tan2010enhanced}.
\end{itemize}

\begin{githubquote}
    \textbf{Observation 4.} The three collaborative strategies (CoLA$^{\intercal}$, CoLA$^{\dagger}$ and CoLA$^{\ddagger}$) have significantly different energy consumption, and more refined collaborative strategies are worth exploring.
\end{githubquote}

We analyze the energy consumption of three collaborative strategies in CoLA (CoLA$^{\intercal}$, CoLA$^{\dagger}$, and CoLA$^{\ddagger}$) to validate the low energy consumption of our experimental setup. Our experiments are conducted on a GPU infrastructure powered by two NVIDIA A800 80GB GPUs and an Intel(R) Xeon(R) Platinum 8358 CPU @ 2.60GHz. We use CodeCarbon~\cite{patterson2021carbon} to record the energy consumption of CoLA$^{\intercal}$, CoLA$^{\dagger}$, and CoLA$^{\ddagger}$ based on Llama-3.1-8B across different single domains, with a random seed of 42. As shown in Figure~\ref{fig:energy}, we observe the following: (1) The energy consumption of the three collaborative strategies—CoLA$^{\intercal}$, CoLA$^{\dagger}$, and CoLA$^{\ddagger}$—differs significantly, representing high, low, and medium configurations, respectively, which are suitable for different real-world application scenarios to meet diverse user needs. (2) Compared to the energy consumption results from HydraLoRA~\cite{tian2024hydralora}, our experiment consumes less than 1/10th of the energy. This is not only due to our sampling a smaller number of samples, which aligns with the scarcity of samples in real-world scenarios and presents a significant challenge for the performance of current PEFT methods, but also due to our experimental setup. We convert the model's generation evaluation into a classification evaluation, resulting in fewer tokens (the model's output is often just a capital letter). More details on our evaluation method can be found in Appendix~\ref{apx: details}.

\begin{table}[]
\resizebox{\columnwidth}{!}{%
\begin{tabular}{@{}c|c|c|c|c|c|c@{}}
\toprule[1.5pt]
\textbf{Domain} & \textbf{Generality} & \textbf{Law} & \textbf{Medicine} & \textbf{Math} & \textbf{Finance} & \textbf{Multi-tasking} \\ \midrule
\textbf{CoLA$^{\dagger}$}  & 42.26               & 26.27        & 41.24             & 45.96         & 37.51            & 39.26                  \\
\textbf{CoLA$^{\widehat{\dagger}}$}  & 52.26               & 31.02        & 48.15             & 54.89         & 46.32            & 40.64                  \\ \bottomrule[1.5pt]
\end{tabular}%
}
\vspace{-0.2cm}
\caption{Performance comparison of two different random collaborative strategies based on Llama-3.1-8B.}
\label{tab:enhance}
\end{table}

However, we also observe that the CoLA$^{\dagger}$ collaborative strategy performs poorly overall in our experiments, as it violates the principle identified in Observation 3. In contrast, we introduce another random collaborative strategy, CoLA$^{\widehat{\dagger}}$, which combines each matrix $B$ with a random matrix $A$. We fine-tune Llama-3.1-8B on multiple different single-domain and multi-domain tasks, as shown in Table~\ref{tab:enhance}. From the table, we find that CoLA$^{\widehat{\dagger}}$ consistently outperforms CoLA$^{\dagger}$, which demonstrates the universality and effectiveness of Observation 3. Furthermore, due to space limitations, more refined collaborative strategies that align with the principles we discover remain to be explored. For example, as shown in Figure~\ref{fig:cola}, where matrix $A$ and $B$ form a bipartite graph, this graph has many interesting properties (\textit{e.g.}, maximum matching), making it a promising avenue for future research.

\section{Conclusion}
\label{sec: conclusion}

In conclusion, this paper introduces CoLA, a flexible LoRA architecture designed to address the limitations of current parameter-efficient fine-tuning methods, particularly in scenarios with scarce data. By decoupling the rigid numerical relationship between matrices $A$ and $B$, CoLA enables more effective collaboration through three various strategies. Through extensive experimentation on multiple Llama models, we demonstrate that CoLA significantly improves generalization and robustness, especially in resource-constrained environments.

\section{Limitations}
\label{sec: limitation}

Our experiment involves multiple knowledge domains: generality, law, medicine, math, finance, and mixed domains (multi-tasking). However, we do not validate the proposed CoLA method in the code domain, which is related to the consistent evaluation approach we adopted (also recommended by the LlamaFactory framework), as shown in Appendix~\ref{apx: details}. Additionally, code-related questions are challenging to transform into multiple-choice questions. We are actively seeking a simple yet non-trivial way to add to the model's input tokens. Furthermore, the more refined collaborative strategy in the CoLA method is worth exploring. For example, as shown in Figure~\ref{fig:cola}, a bipartite graph is formed between matrices $A$ and $B$. Investigating the maximum matching of such a bipartite graph and identifying appropriate scenarios is promising, and we leave this as future work.

\section*{Acknowledgements}

This work was supported by  the "Pioneer" and "Leading Goose" R\&D Program of Zhejiang under Grant No. 2025C02022.

\bibliography{acl2025}

\begin{thebibliography}{63}
\providecommand{\natexlab}[1]{#1}

\bibitem[{Agiza et~al.(2024)Agiza, Neseem, and Reda}]{agiza2024mtlora}
Ahmed Agiza, Marina Neseem, and Sherief Reda. 2024.
\newblock Mtlora: Low-rank adaptation approach for efficient multi-task
  learning.
\newblock In \emph{Proceedings of the IEEE/CVF Conference on Computer Vision
  and Pattern Recognition}, pages 16196--16205.

\bibitem[{Alignment-Lab-AI(2024)}]{Lawyer-Instruct}
Alignment-Lab-AI. 2024.
\newblock Lawyer-instruct.

\bibitem[{Araci(2019)}]{araci2019finbert}
D~Araci. 2019.
\newblock Finbert: Financial sentiment analysis with pre-trained language
  models.
\newblock \emph{arXiv preprint arXiv:1908.10063}.

\bibitem[{bench authors(2023)}]{srivastava2023beyond}
BIG bench authors. 2023.
\newblock \href {https://openreview.net/forum?id=uyTL5Bvosj} {Beyond the
  imitation game: Quantifying and extrapolating the capabilities of language
  models}.
\newblock \emph{Transactions on Machine Learning Research}.

\bibitem[{Chalkidis et~al.(2020)Chalkidis, Fergadiotis, Malakasiotis, Aletras,
  and Androutsopoulos}]{chalkidis2020legal}
Ilias Chalkidis, Manos Fergadiotis, Prodromos Malakasiotis, Nikolaos Aletras,
  and Ion Androutsopoulos. 2020.
\newblock Legal-bert: The muppets straight out of law school.
\newblock \emph{arXiv preprint arXiv:2010.02559}.

\bibitem[{Chalkidis et~al.(2023)Chalkidis, Garneau, Goanta, Katz, and
  S{\o}gaard}]{legal_lama}
Ilias Chalkidis, Nicolas Garneau, Catalina Goanta, Daniel Katz, and Anders
  S{\o}gaard. 2023.
\newblock \href {https://aclanthology.org/2023.acl-long.865} {{L}e{XF}iles and
  {L}egal{LAMA}: Facilitating {E}nglish multinational legal language model
  development}.
\newblock In \emph{Proceedings of the 61st Annual Meeting of the Association
  for Computational Linguistics (Volume 1: Long Papers)}, pages 15513--15535,
  Toronto, Canada. Association for Computational Linguistics.

\bibitem[{Cobbe et~al.(2021)Cobbe, Kosaraju, Bavarian, Chen, Jun, Kaiser,
  Plappert, Tworek, Hilton, Nakano, Hesse, and Schulman}]{gsm8k}
Karl Cobbe, Vineet Kosaraju, Mohammad Bavarian, Mark Chen, Heewoo Jun, Lukasz
  Kaiser, Matthias Plappert, Jerry Tworek, Jacob Hilton, Reiichiro Nakano,
  Christopher Hesse, and John Schulman. 2021.
\newblock Training verifiers to solve math word problems.
\newblock \emph{arXiv preprint arXiv:2110.14168}.

\bibitem[{Conover et~al.(2023)Conover, Hayes, Mathur, Xie, Wan, Shah, Ghodsi,
  Wendell, Zaharia, and Xin}]{conover2023free}
Mike Conover, Matt Hayes, Ankit Mathur, Jianwei Xie, Jun Wan, Sam Shah, Ali
  Ghodsi, Patrick Wendell, Matei Zaharia, and Reynold Xin. 2023.
\newblock Free dolly: Introducing the world’s first truly open
  instruction-tuned llm.
\newblock \emph{Company Blog of Databricks}.

\bibitem[{Dubey et~al.(2024)Dubey, Jauhri, Pandey, Kadian, Al-Dahle, Letman,
  Mathur, Schelten, Yang, Fan et~al.}]{dubey2024llama}
Abhimanyu Dubey, Abhinav Jauhri, Abhinav Pandey, Abhishek Kadian, Ahmad
  Al-Dahle, Aiesha Letman, Akhil Mathur, Alan Schelten, Amy Yang, Angela Fan,
  et~al. 2024.
\newblock The llama 3 herd of models.
\newblock \emph{arXiv preprint arXiv:2407.21783}.

\bibitem[{Feng et~al.(2024)Feng, Hao, Zhang, Han, and Wang}]{feng2024mixture}
Wenfeng Feng, Chuzhan Hao, Yuewei Zhang, Yu~Han, and Hao Wang. 2024.
\newblock Mixture-of-loras: An efficient multitask tuning for large language
  models.
\newblock \emph{arXiv preprint arXiv:2403.03432}.

\bibitem[{Gao et~al.(2024)Gao, Chen, Rao, Sun, Liu, Peng, Zhang, Guo, Yang, and
  Subrahmanian}]{gao2024higher}
Chongyang Gao, Kezhen Chen, Jinmeng Rao, Baochen Sun, Ruibo Liu, Daiyi Peng,
  Yawen Zhang, Xiaoyuan Guo, Jie Yang, and VS~Subrahmanian. 2024.
\newblock Higher layers need more lora experts.
\newblock \emph{arXiv preprint arXiv:2402.08562}.

\bibitem[{Hayou et~al.(2024)Hayou, Ghosh, and Yu}]{hayou2024impact}
Soufiane Hayou, Nikhil Ghosh, and Bin Yu. 2024.
\newblock The impact of initialization on lora finetuning dynamics.
\newblock \emph{arXiv preprint arXiv:2406.08447}.

\bibitem[{Hendrycks et~al.(2020)Hendrycks, Burns, Basart, Zou, Mazeika, Song,
  and Steinhardt}]{hendrycks2020measuring}
Dan Hendrycks, Collin Burns, Steven Basart, Andy Zou, Mantas Mazeika, Dawn
  Song, and Jacob Steinhardt. 2020.
\newblock Measuring massive multitask language understanding.
\newblock \emph{arXiv preprint arXiv:2009.03300}.

\bibitem[{Hendrycks et~al.(2021)Hendrycks, Burns, Basart, Zou, Mazeika, Song,
  and Steinhardt}]{hendryckstest2021}
Dan Hendrycks, Collin Burns, Steven Basart, Andy Zou, Mantas Mazeika, Dawn
  Song, and Jacob Steinhardt. 2021.
\newblock Measuring massive multitask language understanding.
\newblock \emph{Proceedings of the International Conference on Learning
  Representations (ICLR)}.

\bibitem[{Hernandez et~al.(2021)Hernandez, Kaplan, Henighan, and
  McCandlish}]{hernandez2021scaling}
Danny Hernandez, Jared Kaplan, Tom Henighan, and Sam McCandlish. 2021.
\newblock Scaling laws for transfer.
\newblock \emph{arXiv preprint arXiv:2102.01293}.

\bibitem[{Hinton et~al.(2006)Hinton, Osindero, and Teh}]{hinton2006fast}
Geoffrey~E Hinton, Simon Osindero, and Yee-Whye Teh. 2006.
\newblock A fast learning algorithm for deep belief nets.
\newblock \emph{Neural computation}, 18(7):1527--1554.

\bibitem[{Hu et~al.(2021)Hu, Shen, Wallis, Allen-Zhu, Li, Wang, Wang, and
  Chen}]{hu2021lora}
Edward~J Hu, Yelong Shen, Phillip Wallis, Zeyuan Allen-Zhu, Yuanzhi Li, Shean
  Wang, Lu~Wang, and Weizhu Chen. 2021.
\newblock Lora: Low-rank adaptation of large language models.
\newblock \emph{arXiv preprint arXiv:2106.09685}.

\bibitem[{Kaplan et~al.(2020)Kaplan, McCandlish, Henighan, Brown, Chess, Child,
  Gray, Radford, Wu, and Amodei}]{kaplan2020scaling}
Jared Kaplan, Sam McCandlish, Tom Henighan, Tom~B Brown, Benjamin Chess, Rewon
  Child, Scott Gray, Alec Radford, Jeffrey Wu, and Dario Amodei. 2020.
\newblock Scaling laws for neural language models.
\newblock \emph{arXiv preprint arXiv:2001.08361}.

\bibitem[{Ke et~al.(2025)Ke, Wang, Wang, Liu, Nie, Li, and
  Li}]{ke2025unveiling}
Wenjun Ke, Jiahao Wang, Peng Wang, Jiajun Liu, Dong Nie, Guozheng Li, and
  Yining Li. 2025.
\newblock Unveiling lora intrinsic ranks via salience analysis.
\newblock \emph{Advances in Neural Information Processing Systems},
  37:131575--131595.

\bibitem[{LeCun et~al.(1998)LeCun, Bottou, Bengio, and
  Haffner}]{lecun1998gradient}
Yann LeCun, L{\'e}on Bottou, Yoshua Bengio, and Patrick Haffner. 1998.
\newblock Gradient-based learning applied to document recognition.
\newblock \emph{Proceedings of the IEEE}, 86(11):2278--2324.

\bibitem[{Lester et~al.(2021{\natexlab{a}})Lester, Al-Rfou, and
  Constant}]{lester2021power}
Brian Lester, Rami Al-Rfou, and Noah Constant. 2021{\natexlab{a}}.
\newblock The power of scale for parameter-efficient prompt tuning.
\newblock \emph{arXiv preprint arXiv:2104.08691}.

\bibitem[{Lester et~al.(2021{\natexlab{b}})Lester, Al{-}Rfou, and
  Constant}]{LesterAC21}
Brian Lester, Rami Al{-}Rfou, and Noah Constant. 2021{\natexlab{b}}.
\newblock \href {https://doi.org/10.18653/V1/2021.EMNLP-MAIN.243} {The power of
  scale for parameter-efficient prompt tuning}.
\newblock In \emph{Proceedings of the 2021 Conference on Empirical Methods in
  Natural Language Processing, {EMNLP} 2021, Virtual Event / Punta Cana,
  Dominican Republic, 7-11 November, 2021}, pages 3045--3059. Association for
  Computational Linguistics.

\bibitem[{Li et~al.(2025)Li, Zhan, Fu, Zhang, Kuang, Li, Zhao, Wu, and
  Wu}]{li2025mergenet}
Kunxi Li, Tianyu Zhan, Kairui Fu, Shengyu Zhang, Kun Kuang, Jiwei Li, Zhou
  Zhao, Fan Wu, and Fei Wu. 2025.
\newblock Mergenet: Knowledge migration across heterogeneous models, tasks, and
  modalities.
\newblock In \emph{Proceedings of the AAAI Conference on Artificial
  Intelligence}, volume~39, pages 4824--4832.

\bibitem[{Li et~al.(2023)Li, Li, Zhang, Dan, Jiang, and Zhang}]{chatdoctor}
Yunxiang Li, Zihan Li, Kai Zhang, Ruilong Dan, Steve Jiang, and You Zhang.
  2023.
\newblock Chatdoctor: A medical chat model fine-tuned on a large language model
  meta-ai (llama) using medical domain knowledge.
\newblock \emph{Cureus}, 15(6).

\bibitem[{Lian et~al.(2023)Lian, Goodson, Pentland et~al.}]{lian2023openorca}
W~Lian, B~Goodson, E~Pentland, et~al. 2023.
\newblock Openorca: An open dataset of gpt augmented flan reasoning traces.

\bibitem[{Lingam et~al.(2024)Lingam, Tejaswi, Vavre, Shetty, Gudur, Ghosh,
  Dimakis, Choi, Bojchevski, and Sanghavi}]{lingam2024svft}
Vijay Lingam, Atula Tejaswi, Aditya Vavre, Aneesh Shetty, Gautham~Krishna
  Gudur, Joydeep Ghosh, Alex Dimakis, Eunsol Choi, Aleksandar Bojchevski, and
  Sujay Sanghavi. 2024.
\newblock Svft: Parameter-efficient fine-tuning with singular vectors.
\newblock \emph{arXiv preprint arXiv:2405.19597}.

\bibitem[{Liu et~al.(2022)Liu, Tam, Muqeeth, Mohta, Huang, Bansal, and
  Raffel}]{IA3}
Haokun Liu, Derek Tam, Mohammed Muqeeth, Jay Mohta, Tenghao Huang, Mohit
  Bansal, and Colin Raffel. 2022.
\newblock \href
  {http://papers.nips.cc/paper\_files/paper/2022/hash/0cde695b83bd186c1fd456302888454c-Abstract-Conference.html}
  {Few-shot parameter-efficient fine-tuning is better and cheaper than
  in-context learning}.
\newblock In \emph{Advances in Neural Information Processing Systems 35: Annual
  Conference on Neural Information Processing Systems 2022, NeurIPS 2022, New
  Orleans, LA, USA, November 28 - December 9, 2022}.

\bibitem[{Liu et~al.(2024{\natexlab{a}})Liu, Wu, Zhao, Zhu, Xu, Tian, and
  Zheng}]{liu2024moe}
Qidong Liu, Xian Wu, Xiangyu Zhao, Yuanshao Zhu, Derong Xu, Feng Tian, and
  Yefeng Zheng. 2024{\natexlab{a}}.
\newblock When moe meets llms: Parameter efficient fine-tuning for multi-task
  medical applications.
\newblock In \emph{Proceedings of the 47th International ACM SIGIR Conference
  on Research and Development in Information Retrieval}, pages 1104--1114.

\bibitem[{Liu et~al.(2024{\natexlab{b}})Liu, Wang, Yin, Molchanov, Wang, Cheng,
  and Chen}]{liu2024dora}
Shih-Yang Liu, Chien-Yi Wang, Hongxu Yin, Pavlo Molchanov, Yu-Chiang~Frank
  Wang, Kwang-Ting Cheng, and Min-Hung Chen. 2024{\natexlab{b}}.
\newblock Dora: Weight-decomposed low-rank adaptation.
\newblock \emph{arXiv preprint arXiv:2402.09353}.

\bibitem[{Liu et~al.(2021)Liu, Zheng, Du, Ding, Qian, Yang, and
  Tang}]{P-tuning}
Xiao Liu, Yanan Zheng, Zhengxiao Du, Ming Ding, Yujie Qian, Zhilin Yang, and
  Jie Tang. 2021.
\newblock \href {https://arxiv.org/abs/2103.10385} {{GPT} understands, too}.
\newblock \emph{CoRR}, abs/2103.10385.

\bibitem[{Liu and Liu(2010)}]{liu2010fusion}
Zhiming Liu and Chengjun Liu. 2010.
\newblock Fusion of color, local spatial and global frequency information for
  face recognition.
\newblock \emph{Pattern Recognition}, 43(8):2882--2890.

\bibitem[{Longpre et~al.(2023)Longpre, Hou, Vu, Webson, Chung, Tay, Zhou, Le,
  Zoph, Wei et~al.}]{longpre2023flan}
Shayne Longpre, Le~Hou, Tu~Vu, Albert Webson, Hyung~Won Chung, Yi~Tay, Denny
  Zhou, Quoc~V Le, Barret Zoph, Jason Wei, et~al. 2023.
\newblock The flan collection: Designing data and methods for effective
  instruction tuning.
\newblock In \emph{International Conference on Machine Learning}, pages
  22631--22648. PMLR.

\bibitem[{Meng et~al.(2024)Meng, Wang, and Zhang}]{meng2024pissa}
Fanxu Meng, Zhaohui Wang, and Muhan Zhang. 2024.
\newblock Pissa: Principal singular values and singular vectors adaptation of
  large language models.
\newblock \emph{arXiv preprint arXiv:2404.02948}.

\bibitem[{Mukherjee et~al.(2023)Mukherjee, Mitra, Jawahar, Agarwal, Palangi,
  and Awadallah}]{mukherjee2023orca}
Subhabrata Mukherjee, Arindam Mitra, Ganesh Jawahar, Sahaj Agarwal, Hamid
  Palangi, and Ahmed Awadallah. 2023.
\newblock Orca: Progressive learning from complex explanation traces of gpt-4.
\newblock \emph{arXiv preprint arXiv:2306.02707}.

\bibitem[{Ouyang et~al.(2022)Ouyang, Wu, Jiang, Almeida, Wainwright, Mishkin,
  Zhang, Agarwal, Slama, Ray et~al.}]{ouyang2022training}
Long Ouyang, Jeffrey Wu, Xu~Jiang, Diogo Almeida, Carroll Wainwright, Pamela
  Mishkin, Chong Zhang, Sandhini Agarwal, Katarina Slama, Alex Ray, et~al.
  2022.
\newblock Training language models to follow instructions with human feedback.
\newblock \emph{Advances in neural information processing systems},
  35:27730--27744.

\bibitem[{Patel et~al.(2024)Patel, Timsina, Raut, Freeman, levin, Nadkarni,
  Glicksberg, and Klang}]{patel2024exploring}
Dhavalkumar Patel, Prem Timsina, Ganesh Raut, Robert Freeman, Matthew~A levin,
  Girish~N Nadkarni, Benjamin~S Glicksberg, and Eyal Klang. 2024.
\newblock Exploring temperature effects on large language models across various
  clinical tasks.
\newblock \emph{medRxiv}, pages 2024--07.

\bibitem[{Patterson et~al.(2021)Patterson, Gonzalez, Le, Liang, Munguia,
  Rothchild, So, Texier, and Dean}]{patterson2021carbon}
David Patterson, Joseph Gonzalez, Quoc Le, Chen Liang, Lluis-Miquel Munguia,
  Daniel Rothchild, David So, Maud Texier, and Jeff Dean. 2021.
\newblock Carbon emissions and large neural network training.
\newblock \emph{arXiv preprint arXiv:2104.10350}.

\bibitem[{Pecher et~al.(2024)Pecher, Srba, and Bielikova}]{pecher2024fine}
Branislav Pecher, Ivan Srba, and Maria Bielikova. 2024.
\newblock Fine-tuning, prompting, in-context learning and instruction-tuning:
  How many labelled samples do we need?
\newblock \emph{arXiv preprint arXiv:2402.12819}.

\bibitem[{Peng et~al.(2021)Peng, Yuan, Gao, and Tang}]{peng2021mathbert}
Shuai Peng, Ke~Yuan, Liangcai Gao, and Zhi Tang. 2021.
\newblock Mathbert: A pre-trained model for mathematical formula understanding.
\newblock \emph{arXiv preprint arXiv:2105.00377}.

\bibitem[{Qin et~al.(2021)Qin, Wang, Su, Lin, Ding, Yi, Chen, Liu, Li, Hou
  et~al.}]{qin2021exploring}
Yujia Qin, Xiaozhi Wang, Yusheng Su, Yankai Lin, Ning Ding, Jing Yi, Weize
  Chen, Zhiyuan Liu, Juanzi Li, Lei Hou, et~al. 2021.
\newblock Exploring universal intrinsic task subspace via prompt tuning.
\newblock \emph{arXiv preprint arXiv:2110.07867}.

\bibitem[{Rasmy et~al.(2021)Rasmy, Xiang, Xie, Tao, and Zhi}]{rasmy2021med}
Laila Rasmy, Yang Xiang, Ziqian Xie, Cui Tao, and Degui Zhi. 2021.
\newblock Med-bert: pretrained contextualized embeddings on large-scale
  structured electronic health records for disease prediction.
\newblock \emph{NPJ digital medicine}, 4(1):86.

\bibitem[{Renze and Guven(2024)}]{renze2024effect}
Matthew Renze and Erhan Guven. 2024.
\newblock The effect of sampling temperature on problem solving in large
  language models.
\newblock \emph{arXiv preprint arXiv:2402.05201}.

\bibitem[{Riesenhuber and Poggio(1999)}]{riesenhuber1999hierarchical}
Maximilian Riesenhuber and Tomaso Poggio. 1999.
\newblock Hierarchical models of object recognition in cortex.
\newblock \emph{Nature neuroscience}, 2(11):1019--1025.

\bibitem[{Sch{\"a}fer et~al.(2024)Sch{\"a}fer, Nicke, H{\"o}fener, Lange,
  Merhof, Feuerhake, Schulz, Lotz, and Kiessling}]{schafer2024overcoming}
Raphael Sch{\"a}fer, Till Nicke, Henning H{\"o}fener, Annkristin Lange, Dorit
  Merhof, Friedrich Feuerhake, Volkmar Schulz, Johannes Lotz, and Fabian
  Kiessling. 2024.
\newblock Overcoming data scarcity in biomedical imaging with a foundational
  multi-task model.
\newblock \emph{Nature Computational Science}, 4(7):495--509.

\bibitem[{Srivastava et~al.(2014)Srivastava, Hinton, Krizhevsky, Sutskever, and
  Salakhutdinov}]{srivastava2014dropout}
Nitish Srivastava, Geoffrey Hinton, Alex Krizhevsky, Ilya Sutskever, and Ruslan
  Salakhutdinov. 2014.
\newblock Dropout: a simple way to prevent neural networks from overfitting.
\newblock \emph{The journal of machine learning research}, 15(1):1929--1958.

\bibitem[{Suzgun et~al.(2022)Suzgun, Scales, Sch{\"a}rli, Gehrmann, Tay, Chung,
  Chowdhery, Le, Chi, Zhou, , and Wei}]{suzgun2022challenging}
Mirac Suzgun, Nathan Scales, Nathanael Sch{\"a}rli, Sebastian Gehrmann, Yi~Tay,
  Hyung~Won Chung, Aakanksha Chowdhery, Quoc~V Le, Ed~H Chi, Denny Zhou, , and
  Jason Wei. 2022.
\newblock Challenging big-bench tasks and whether chain-of-thought can solve
  them.
\newblock \emph{arXiv preprint arXiv:2210.09261}.

\bibitem[{Tan and Triggs(2010)}]{tan2010enhanced}
Xiaoyang Tan and Bill Triggs. 2010.
\newblock Enhanced local texture feature sets for face recognition under
  difficult lighting conditions.
\newblock \emph{IEEE transactions on image processing}, 19(6):1635--1650.

\bibitem[{Tian et~al.(2024)Tian, Shi, Guo, Li, and Xu}]{tian2024hydralora}
Chunlin Tian, Zhan Shi, Zhijiang Guo, Li~Li, and Chengzhong Xu. 2024.
\newblock Hydralora: An asymmetric lora architecture for efficient fine-tuning.
\newblock \emph{arXiv preprint arXiv:2404.19245}.

\bibitem[{Van(2023)}]{van2023mitigating}
Hoang Van. 2023.
\newblock Mitigating data scarcity for large language models.
\newblock \emph{arXiv preprint arXiv:2302.01806}.

\bibitem[{Wang et~al.(2023)Wang, Yang, and Wang}]{wang2023fingptbenchmark}
Neng Wang, Hongyang Yang, and Christina~Dan Wang. 2023.
\newblock Fingpt: Instruction tuning benchmark for open-source large language
  models in financial datasets.
\newblock \emph{NeurIPS Workshop on Instruction Tuning and Instruction
  Following}.

\bibitem[{Wang et~al.(2024{\natexlab{a}})Wang, Yu, and Li}]{wang2024lora}
Shaowen Wang, Linxi Yu, and Jian Li. 2024{\natexlab{a}}.
\newblock Lora-ga: Low-rank adaptation with gradient approximation.
\newblock \emph{arXiv preprint arXiv:2407.05000}.

\bibitem[{Wang et~al.(2024{\natexlab{b}})Wang, Liang, He, Wang, and
  Tan}]{wang2024lorapro}
Zhengbo Wang, Jian Liang, Ran He, Zilei Wang, and Tieniu Tan.
  2024{\natexlab{b}}.
\newblock Lora-pro: Are low-rank adapters properly optimized?
\newblock \emph{arXiv preprint arXiv:2407.18242}.

\bibitem[{Yang et~al.(2024{\natexlab{a}})Yang, Muhtar, Shen, Zhan, Liu, Wang,
  Sun, Deng, Sun, Zhang et~al.}]{yang2024mtl}
Yaming Yang, Dilxat Muhtar, Yelong Shen, Yuefeng Zhan, Jianfeng Liu, Yujing
  Wang, Hao Sun, Denvy Deng, Feng Sun, Qi~Zhang, et~al. 2024{\natexlab{a}}.
\newblock Mtl-lora: Low-rank adaptation for multi-task learning.
\newblock \emph{arXiv preprint arXiv:2410.09437}.

\bibitem[{Yang et~al.(2024{\natexlab{b}})Yang, Li, Zhou, Song, Wu, Nie, and
  Ghanem}]{yang2024corda}
Yibo Yang, Xiaojie Li, Zhongzhu Zhou, Shuaiwen~Leon Song, Jianlong Wu, Liqiang
  Nie, and Bernard Ghanem. 2024{\natexlab{b}}.
\newblock Corda: Context-oriented decomposition adaptation of large language
  models.
\newblock \emph{arXiv preprint arXiv:2406.05223}.

\bibitem[{Zhai et~al.(2022)Zhai, Kolesnikov, Houlsby, and
  Beyer}]{zhai2022scaling}
Xiaohua Zhai, Alexander Kolesnikov, Neil Houlsby, and Lucas Beyer. 2022.
\newblock Scaling vision transformers.
\newblock In \emph{Proceedings of the IEEE/CVF conference on computer vision
  and pattern recognition}, pages 12104--12113.

\bibitem[{Zhang and Pilanci(2024)}]{zhang2024spectral}
Fangzhao Zhang and Mert Pilanci. 2024.
\newblock Spectral adapter: Fine-tuning in spectral space.
\newblock \emph{arXiv preprint arXiv:2405.13952}.

\bibitem[{Zhao et~al.(2003)Zhao, Chellappa, Phillips, and
  Rosenfeld}]{zhao2003face}
Wenyi Zhao, Rama Chellappa, P~Jonathon Phillips, and Azriel Rosenfeld. 2003.
\newblock Face recognition: A literature survey.
\newblock \emph{ACM computing surveys (CSUR)}, 35(4):399--458.

\bibitem[{Zheng et~al.(2024)Zheng, Zhang, Zhang, Ye, Luo, Feng, and
  Ma}]{zheng2024llamafactory}
Yaowei Zheng, Richong Zhang, Junhao Zhang, Yanhan Ye, Zheyan Luo, Zhangchi
  Feng, and Yongqiang Ma. 2024.
\newblock \href {http://arxiv.org/abs/2403.13372} {Llamafactory: Unified
  efficient fine-tuning of 100+ language models}.
\newblock In \emph{Proceedings of the 62nd Annual Meeting of the Association
  for Computational Linguistics (Volume 3: System Demonstrations)}, Bangkok,
  Thailand. Association for Computational Linguistics.

\bibitem[{Zhou et~al.(2025{\natexlab{a}})Zhou, Han, and
  Chen}]{zhou2025revisiting}
Yiyun Zhou, Wenkang Han, and Jingyuan Chen. 2025{\natexlab{a}}.
\newblock Revisiting applicable and comprehensive knowledge tracing in
  large-scale data.
\newblock \emph{arXiv preprint arXiv:2501.14256}.

\bibitem[{Zhou et~al.(2024)Zhou, Lv, Zhang, and Chen}]{zhou2024cuffkt}
Yiyun Zhou, Zheqi Lv, Shengyu Zhang, and Jingyuan Chen. 2024.
\newblock \href {https://openreview.net/forum?id=UVaPEthRKx} {Cuff-{KT}:
  Tackling learners' real-time learning pattern adjustment via tuning-free
  knowledge state-guided model updating}.

\bibitem[{Zhou et~al.(2025{\natexlab{b}})Zhou, Lv, Zhang, and
  Chen}]{zhou2025disentangled}
Yiyun Zhou, Zheqi Lv, Shengyu Zhang, and Jingyuan Chen. 2025{\natexlab{b}}.
\newblock Disentangled knowledge tracing for alleviating cognitive bias.
\newblock In \emph{Proceedings of the ACM on Web Conference 2025}, pages
  2633--2645.

\bibitem[{Zhu et~al.(2024{\natexlab{a}})Zhu, Greenewald, Nadjahi, Borde,
  Gabrielsson, Choshen, Ghassemi, Yurochkin, and Solomon}]{zhu2024asymmetry}
Jiacheng Zhu, Kristjan Greenewald, Kimia Nadjahi, Haitz S{\'a}ez de~Oc{\'a}riz
  Borde, Rickard~Br{\"u}el Gabrielsson, Leshem Choshen, Marzyeh Ghassemi,
  Mikhail Yurochkin, and Justin Solomon. 2024{\natexlab{a}}.
\newblock Asymmetry in low-rank adapters of foundation models.
\newblock \emph{arXiv preprint arXiv:2402.16842}.

\bibitem[{Zhu et~al.(2024{\natexlab{b}})Zhu, Li, Li, Zhao, Jin, and
  Mei}]{zhu2024hot}
Yuqi Zhu, Jia Li, Ge~Li, YunFei Zhao, Zhi Jin, and Hong Mei.
  2024{\natexlab{b}}.
\newblock Hot or cold? adaptive temperature sampling for code generation with
  large language models.
\newblock In \emph{Proceedings of the AAAI Conference on Artificial
  Intelligence}, volume~38, pages 437--445.

\end{thebibliography}

\clearpage

\appendix

\section{Datasets}
\label{apx: datasets}

In our experiment, a total of 8 datasets are used for fine-tuning, and their descriptions are as follows.

\begin{itemize}
    \item \textbf{databricks-dolly-15k:} databricks-dolly-15k is an open-source dataset with 15,000 high-quality, human-generated prompt/response pairs, created by over 5,000 Databricks employees in March and April 2023. It is designed for instruction tuning LLMs and includes expressive and diverse examples across various tasks such as brainstorming, content generation, classification, summarization, information extraction, closed QA, and open QA. Inspired by the behavioral categories in the InstructGPT~\cite{ouyang2022training}, this dataset supports a wide range of instruction-following applications.
    \item  \textbf{Lawyer-Instruct:} Lawyer-Instruct is an English-based conversational dataset derived from the original LawyerChat dataset. It features legal dialogue scenarios restructured into a format with clear instructions, inputs, and expected outputs. This redesigned format makes it particularly suitable for training supervised dialogue models.
    \item \textbf{US-Terms:} LegalLAMA is a comprehensive benchmark suite consisting of 8 sub-tasks, designed to evaluate the extent of legal knowledge acquired by PLMs during pre-training. US-Terms is one of these sub-tasks.
    \item \textbf{GenMedGPT-5k, clinic-10k:} GenMedGPT-5k and icliniq-10k both originate from ChatDoctor. GenMedGPT-5k contains a dataset of over 5,000 GPT-generated doctor-patient dialogues, while icliniq-10k includes a dataset of over 10,000 patient-doctor dialogues.
    \item \textbf{GSM8k:} GSM8K (Grade School Math 8K) is a dataset of 8.5K high-quality, linguistically diverse grade school math word problems. The dataset is designed to support the task of question answering on basic mathematical problems that require multi-step reasoning. We use the option-filled prompt shown in Appendix~\ref{apx: details} to convert it into multiple-choice questions.
    \item \textbf{fingpt-fineval:} fingpt-fineval comes from FinGPT and includes Chinese multiple-choice questions instructions. However, the Llama model family does not support the Chinese language, so we used Google Cloud Translation\footnote{\url{https://cloud.google.com/translation-hub}} to translate it into English.
    \item \textbf{OpenOrca:} OpenOrca is an instruction-tuning dataset, with 2.91M samples derived from augmented FLAN~\cite{longpre2023flan}. It includes around 1M GPT-4 completions and 3.2M GPT-3.5 completions, organized as per ORCA's distribution~\cite{mukherjee2023orca}, aimed at training and evaluation in natural language processing tasks.
\end{itemize}

It is worth noting that we do not fine-tune the entire aforementioned datasets, but instead randomly select 1,000 samples with a random seed of 42, based on the following considerations: \texttt{(i)}\textbf{ In real-world scenarios, fine-tuning samples are scarce~\cite{van2023mitigating, schafer2024overcoming, zhou2024cuffkt, pecher2024fine}.} Data collection and labeling are costly, and high-quality samples are even harder to obtain for specific tasks. The scarcity of samples can easily lead to model overfitting, which poses a significant challenge for all PEFT methods. \texttt{(ii)} \textbf{Limited resources.} All resources are utilized on two NVIDIA A800-SXM4 (80G) GPUs, but most research institutions do not even have such configurations. We provide additional experiments with larger sample sizes in Appendix~\ref{apx: results} to demonstrate the robustness of the proposed CoLA.

In addition, the descriptions of the MMLU and BBH datasets used in our experiments are as follows.

\begin{itemize}
    \item \textbf{MMLU:} MMLU (Massive Multitask Language Understanding) is a benchmark designed to evaluate models in zero-shot and few-shot settings, covering 57 subjects across diverse fields like STEM, humanities, and social sciences. It tests both world knowledge and problem-solving ability, ranging from basic to advanced levels, and helps identify models' blind spots. The test challenges models to demonstrate extensive knowledge across multiple domains.
    \item \textbf{BBH:} BBH (BIG-Bench Hard) is a challenging subset of the BIG-Bench~\cite{srivastava2023beyond}, developed by Google and Stanford. It consists of 23 tasks that require multi-step reasoning, testing large language models' logical and reasoning abilities. 
\end{itemize}

It's noted that we select tasks related to law (international law, jurisprudence, professional law) and medicine (anatomy, clinical knowledge, college medicine, human aging, human sexuality, medical genetics, professional medicine, virology) in MMLU for evaluation in the fields of law and medicine. We use the complete benchmark data for evaluation to ensure the adequacy and comprehensiveness of the experiments.

\section{Baselines}
\label{apx: baselines}

In our experiment, a total of 10 different PEFT methods are used to optimize the recent Llama models, described as follows.

\begin{itemize}
    \item \textbf{Prompt Tuning:} Prompt Tuning introduces task-specific prompts to the input, updating only the prompt parameters while keeping the pretrained model's parameters frozen. It treats all tasks as generation tasks, with prompts being the focus of adaptation.
    \item \textbf{P-Tuning:} P-Tuning introduces trainable prompt embeddings optimized by a prompt encoder, eliminating manual prompt design. It allows prompt tokens to be added anywhere in the input sequence and includes anchor tokens to enhance performance.
    \item \textbf{IA$^{3}$:} IA$^{3}$ improves efficiency by integrating learned vectors into transformer models, reducing trainable parameters while maintaining performance and minimizing inference latency. This PEFT method involves multiplying model activations by three learned vectors, offering a more efficient alternative to LoRA with fewer parameters to update.
    \item \textbf{LoRA:} LoRA is a low-rank decomposition technique that reduces trainable parameters, speeding up fine-tuning and reducing memory usage by inserting trainable low-rank parameters into the original model weights.
    \item \textbf{DoRA:} DoRA decomposes the pre-trained weights into two components—magnitude and direction—to facilitate fine-tuning, using LoRA specifically for directional updates, which effectively minimizes the number of trainable parameters.
    \item \textbf{PiSSA:} PiSSA improves upon LoRA by initializing the adapter with principal singular values and vectors, optimizing the key components while freezing the "noisy" ones. This method leads to faster convergence and better performance than LoRA.
    \item \textbf{HydraLoRA:} HydraLoRA is an asymmetric fine-tuning architecture that effectively identifies and adapts to intrinsic data components, such as sub-domains or diverse tasks. It allocates distinct B matrices for task-specific features, while a shared A matrix integrates global information, enabling efficient parameter utilization and enhanced performance.
    \item \textbf{MOELoRA:} MOELoRA combines the benefits of multi-task learning and parameter-efficient fine-tuning by using multiple experts, each consisting of a pair of low-rank matrices, keeping trainable parameters minimal. The expert modules enable MOELoRA to handle task differences effectively and mitigate the negative impact of data imbalance on performance.
    \item \textbf{MTL-LoRA:} MTL-LoRA enhances low-rank adaptation (LoRA) by adding task-specific parameters, improving multi-task learning. It enables LLMs to adapt to diverse tasks efficiently, using fewer trainable parameters while capturing shared knowledge across tasks in low-dimensional spaces.
    \item \textbf{MoLA:} MoLA for Transformer-based models allows each layer to use a variable number of LoRA experts, with more experts allocated to higher layers, enhancing model effectiveness while maintaining a fixed total number of experts.
\end{itemize}

The PEFT methods related to LoRA mentioned above are used to optimize all linear modules (down\_proj, k\_proj, v\_proj, q\_proj, up\_proj, gate\_proj, o\_proj).

\section{Two Types of Inference Modes}
\label{apx: modes}

Codes~\ref{list: logit} and~\ref{list: text} represent the code for two different inference modes, respectively. As can be observed, Code~\ref{list: logit} is more suitable for classification tasks (\textit{e.g.},  multiple-choice questions), as it is based on the model's logits for inference. When selecting the final answer, the decision is made directly by choosing the option with the highest probability. This mode does not involve a text generation process, so its results are relatively stable and consistent. The evaluation accuracy of this mode is not affected by the randomness of the generation process. In contrast, Code~\ref{list: text} is based on generation tasks (\textit{e.g.}, text generation, question answering, etc.). It generates the entire text to obtain the answer (extracted via regular expressions, with different studies even using different extraction methods), and the quality of the generated output is significantly influenced by sampling strategies (\textit{e.g.}, temperature). A trade-off must be made between diversity and accuracy in the generation process. For reproducibility, we select the first mode recommended by LlamaFactory for all experiments.

\begin{lstlisting}[language=Python, caption={Inference mode with logit.}, label={list: logit}]
@torch.inference_mode()
def batch_inference_logit(self, batch_input: Dict[str, "torch.Tensor"]) -> List[str]:
    # self.choice_inputs: Encoding of the options in the instruction
    logits = self.model(**batch_input).logits
    lengths = torch.sum(batch_input["attention_mask"], dim=-1)
    word_probs = torch.stack([logits[i, lengths[i] - 1] for i in range(len(lengths))], dim=0)
    choice_probs = torch.nn.functional.softmax(word_probs[:, self.choice_inputs], dim=-1).detach()
    return [chr(ord("A") + offset.item()) for offset in torch.argmax(choice_probs, dim=-1)]
\end{lstlisting}

\begin{lstlisting}[language=Python, caption={Inference mode with text.}, label={list: text}]
@torch.inference_mode()
def batch_inference_text(self, batch_input: Dict[str, "torch.Tensor"]) -> List[str]:
    outputs = self.model.generate(
        input_ids=batch_input["input_ids"],
        attention_mask=batch_input["attention_mask"],
        max_length=self.eval_args.max_answer_length,  # Adjustable maximum generated length
        num_beams=1, 
        early_stopping=True
    )
    return [self.tokenizer.decode(output, skip_special_tokens=True) for output in outputs]
\end{lstlisting} 

\section{Experimental Details}
\label{apx: details}

Some fine-tuning datasets (GSM8K and subsets of BBH) have instruction formats that are not based on multiple-choice questions, so they need to be extended through answer options to adapt to classification tasks. We generate additional options using the following option-filling prompt.

\begin{myboxi}[$\bullet$ The Prompt for Option Filling]
Now there is a question about $\{\texttt{subject}\}$ and a correct option. Please fill in the other incorrect options based on the question's context. Note that you should add the incorrect options, not solve the question.\\
---\\
\textbf{Question:} Henry and 3 of his friends order 7 pizzas for lunch. Each pizza is cut into 8 slices. If Henry and his friends want to share the pizzas equally, how many slices can each of them have?\\
\textbf{Correct Option:} A. 14\\
\textbf{Answer:} \\
B. 56\\
C. 8\\
D. 18\\
\\
\textbf{Question:} Farmer Brown has 20 animals on his farm, all either chickens or cows. They have a total of 70 legs, all together. How many of the animals are chickens?\\
\textbf{Correct Option:} C. 5\\
\textbf{Answer:} \\
A. 20\\
B. 15\\
D. 70\\
\\
\textbf{Question:} $\{\texttt{question}\}$\\
\textbf{Correct Option:} $\{\texttt{correct\_option}\}$\\
\textbf{Answer:}\\
$\{\texttt{incorrect\_options}\}$
\end{myboxi}

We use the zero-cost API GLM-4-Flash\footnote{\url{https://bigmodel.cn/dev/activities/free/glm-4-flash}} to accomplish the above task. In addition, the following evaluation template in our experiments is used, as shown in Figure~\ref{fig:temp}.

\begin{figure}[!htb]
\
\centering
\resizebox{0.98\linewidth}{!}{%
\fcolorbox{black}{gray!10}{
\parbox{\linewidth}{
\noindent$\bullet$ \textbf{USER.}

The following are multiple choice questions (with answers) about $\{\texttt{subject}\}$. Please provide only the correct option (one uppercase letter).

$\{\texttt{instruction}\}$

Answer:

\noindent$\bullet$ \textbf{ASSISTANT.} 

$\{\texttt{correct\_option}\}$
}
}
}
\vspace{-0.3cm}
\caption{Evaluation Template}
\vspace{-0.3cm}

\label{fig:temp}
\end{figure}

\begin{table*}[]
\resizebox{\textwidth}{!}{%
\begin{tabular}{@{}cl|c|c|c|c|c|cl|cl|cl|cl|cl|cl|cl|cl|cl|cl|cl@{}}
\toprule[1.5pt]
\multicolumn{2}{c|}{\textbf{Method}}                            & \textbf{Llama-3.2-3B} & \textbf{FFT} & \textbf{Prompt Tuning} & \textbf{P-Tuning} & \textbf{IA$^{3}$} & \multicolumn{2}{c|}{\textbf{LoRA$_{r=8}$}}         & \multicolumn{2}{c|}{\textbf{LoRA$_{r=16}$}}        & \multicolumn{2}{c|}{\textbf{LoRA$_{r=24}$}}        & \multicolumn{2}{c|}{\textbf{LoRA$_{r=32}$}}        & \multicolumn{2}{c|}{\textbf{DoRA}}                  & \multicolumn{2}{c|}{\textbf{PiSSA}}                 & \multicolumn{2}{c|}{\textbf{HydraLoRA}}             & \multicolumn{2}{c|}{\textbf{CoLA}}                            & \multicolumn{2}{c|}{\textbf{CoLA$^{\intercal}$}}                           & \multicolumn{2}{c|}{\textbf{CoLA$^{\dagger}$}}                     & \multicolumn{2}{c}{\textbf{CoLA$^{\ddagger}$}}                     \\ \midrule
\rowcolor[HTML]{EFEFEF} 
\multicolumn{2}{c|}{\cellcolor[HTML]{EFEFEF}\textbf{\#A \big| \#B}} & -                     & -            & -                      & -                 & -                 & \multicolumn{2}{c|}{\cellcolor[HTML]{EFEFEF}1\,\,\,\, \big| \,\,\,\,1}      & \multicolumn{2}{c|}{\cellcolor[HTML]{EFEFEF}1\,\,\,\, \big| \,\,\,\,1}      & \multicolumn{2}{c|}{\cellcolor[HTML]{EFEFEF}1\,\,\,\, \big| \,\,\,\,1}      & \multicolumn{2}{c|}{\cellcolor[HTML]{EFEFEF}1\,\,\,\, \big| \,\,\,\,1}      & \multicolumn{2}{c|}{\cellcolor[HTML]{EFEFEF}1\,\, \big| \,\,1}      & \multicolumn{2}{c|}{\cellcolor[HTML]{EFEFEF}1\,\, \big| \,\,1}      & \multicolumn{2}{c|}{\cellcolor[HTML]{EFEFEF}1\,\,\,\,\,\, \big| \,\,\,\,\,\,3}      & \multicolumn{2}{c|}{\cellcolor[HTML]{EFEFEF}1\,\, \big| \,\,3}                & \multicolumn{2}{c|}{\cellcolor[HTML]{EFEFEF}2\,\, \big| \,\,3}                & \multicolumn{2}{c|}{\cellcolor[HTML]{EFEFEF}2\,\, \big| \,\,3}          & \multicolumn{2}{c}{\cellcolor[HTML]{EFEFEF}2\,\, \big| \,\,3}          \\ \midrule
\rowcolor[HTML]{EFEFEF} 
\multicolumn{2}{c|}{\cellcolor[HTML]{EFEFEF}\textbf{\%Param}}   & -                     & -            & 0.0008                 & 0.0530            & 0.0089            & \multicolumn{2}{c|}{\cellcolor[HTML]{EFEFEF}0.3770} & \multicolumn{2}{c|}{\cellcolor[HTML]{EFEFEF}0.7511} & \multicolumn{2}{c|}{\cellcolor[HTML]{EFEFEF}1.1224} & \multicolumn{2}{c|}{\cellcolor[HTML]{EFEFEF}1.4910} & \multicolumn{2}{c|}{\cellcolor[HTML]{EFEFEF}0.3770} & \multicolumn{2}{c|}{\cellcolor[HTML]{EFEFEF}0.3770} & \multicolumn{2}{c|}{\cellcolor[HTML]{EFEFEF}0.8267} & \multicolumn{2}{c|}{\cellcolor[HTML]{EFEFEF}0.7581}           & \multicolumn{2}{c|}{\cellcolor[HTML]{EFEFEF}0.9406}           & \multicolumn{2}{c|}{\cellcolor[HTML]{EFEFEF}0.9406}     & \multicolumn{2}{c}{\cellcolor[HTML]{EFEFEF}0.9406}     \\ \midrule
\multicolumn{2}{c|}{\textbf{Generality}}                        & 23.01                 & 32.87        & 24.71                  & 25.02             & 24.56             & \multicolumn{2}{c|}{26.28}                          & \multicolumn{2}{c|}{31.25}                          & \multicolumn{2}{c|}{32.88}                          & \multicolumn{2}{c|}{34.90}                          & \multicolumn{2}{c|}{26.38}                          & \multicolumn{2}{c|}{\underline{37.33}}                          & \multicolumn{2}{c|}{25.37}                          & \multicolumn{2}{c|}{\cellcolor[HTML]{CBCEFB}\textbf{45.36**}} & \multicolumn{2}{c|}{\cellcolor[HTML]{CBCEFB}\textbf{46.99**}} & \multicolumn{2}{c|}{\cellcolor[HTML]{CBCEFB}\textbf{26.53}} & \multicolumn{2}{c}{\cellcolor[HTML]{CBCEFB}\textbf{43.15**}} \\
\multicolumn{2}{c|}{\textbf{Law}}                               & 24.73                 & \underline{26.28}        & 23.09                  & 25.69             & 24.94             & \multicolumn{2}{c|}{24.90}                          & \multicolumn{2}{c|}{25.07}                          & \multicolumn{2}{c|}{24.73}                          & \multicolumn{2}{c|}{24.56}                          & \multicolumn{2}{c|}{24.96}                          & \multicolumn{2}{c|}{24.56}                          & \multicolumn{2}{c|}{24.79}                          & \multicolumn{2}{c|}{\cellcolor[HTML]{CBCEFB}\textbf{27.51}} & \multicolumn{2}{c|}{\cellcolor[HTML]{CBCEFB}\textbf{26.85}}   & \multicolumn{2}{c|}{\cellcolor[HTML]{CBCEFB}\textbf{24.89}} & \multicolumn{2}{c}{\cellcolor[HTML]{CBCEFB}\textbf{24.99}} \\
\multicolumn{2}{c|}{\textbf{Medicine}}                          & 24.03                 &         \underline{32.29}     & 23.28                  & 25.39             & 24.24             & \multicolumn{2}{c|}{24.71}                          & \multicolumn{2}{c|}{25.39}                          & \multicolumn{2}{c|}{25.80}                          & \multicolumn{2}{c|}{25.87}                          & \multicolumn{2}{c|}{24.57}                          & \multicolumn{2}{c|}{27.10}                          & \multicolumn{2}{c|}{24.51}                          & \multicolumn{2}{c|}{\cellcolor[HTML]{CBCEFB}\textbf{39.86**}} & \multicolumn{2}{c|}{\cellcolor[HTML]{CBCEFB}\textbf{37.41**}} & \multicolumn{2}{c|}{\cellcolor[HTML]{CBCEFB}\textbf{24.91}} & \multicolumn{2}{c}{\cellcolor[HTML]{CBCEFB}\textbf{33.08*}} \\
\multicolumn{2}{c|}{\textbf{Math}}                              & 24.87                 &      \underline{53.92}        & 25.17                  & 25.09             & 24.79             & \multicolumn{2}{c|}{45.87}                          & \multicolumn{2}{c|}{50.04}                          & \multicolumn{2}{c|}{50.11}                          & \multicolumn{2}{c|}{52.16}                          & \multicolumn{2}{c|}{45.56}                          & \multicolumn{2}{c|}{53.50}                          & \multicolumn{2}{c|}{44.73}                          & \multicolumn{2}{c|}{\cellcolor[HTML]{CBCEFB}\textbf{56.71*}} & \multicolumn{2}{c|}{\cellcolor[HTML]{CBCEFB}\textbf{56.79*}} & \multicolumn{2}{c|}{\cellcolor[HTML]{CBCEFB}\textbf{41.72}} & \multicolumn{2}{c}{\cellcolor[HTML]{CBCEFB}\textbf{52.86}} \\
\multicolumn{2}{c|}{\textbf{Finance}}                           & 26.42                 &        \underline{42.03}      & 24.53                  & 25.66             & 26.78             & \multicolumn{2}{c|}{35.09}                          & \multicolumn{2}{c|}{36.60}                          & \multicolumn{2}{c|}{39.25}                          & \multicolumn{2}{c|}{39.62}                          & \multicolumn{2}{c|}{34.34}                          & \multicolumn{2}{c|}{40.38}                          & \multicolumn{2}{c|}{33.21}                          & \multicolumn{2}{c|}{\cellcolor[HTML]{CBCEFB}\textbf{39.62}}   & \multicolumn{2}{c|}{\cellcolor[HTML]{CBCEFB}\textbf{41.51}} & \multicolumn{2}{c|}{\cellcolor[HTML]{CBCEFB}\textbf{38.93}} & \multicolumn{2}{c}{\cellcolor[HTML]{CBCEFB}\textbf{40.32}} \\ \bottomrule[1.5pt]
\end{tabular}%
}
\vspace{-0.2cm}
\caption{Comparison of 0-shot performance (\%) of different fine-tuning methods based on Llama-3.2-3B across multiple single domains.}
\vspace{-0.3cm}
\label{tab:single_domain_other}
\end{table*}

\begin{table}[t]
\resizebox{\columnwidth}{!}{%
\begin{tabular}{@{}c|c@{}}
\toprule[1.5pt]
\textbf{Hyperparameter}              & \textbf{Setting}                         \\ \midrule
\midrule
Batch Size                  & 8                               \\
\rowcolor{gray!20}
\multicolumn{1}{c|}{Train Epochs}                & \multicolumn{1}{c}{5.0}                             \\
Validation Size             & 0.1                             \\
\rowcolor{gray!20} 
\multicolumn{1}{c|}{Learning Rate}               & \multicolumn{1}{c}{5e-5}                            \\
Cutoff Length               & 1024                            \\
\rowcolor{gray!20}
\multicolumn{1}{c|}{Gradient Accumulation Steps} & \multicolumn{1}{c}{8}                               \\
Random Seed                       & 42,43,44,45,46                  \\
\rowcolor{gray!20} 
\multicolumn{1}{c|}{Scheduler Type}              & \multicolumn{1}{c}{cosine}                          \\
Precision                   & fp16                            \\
\rowcolor{gray!20}
\multicolumn{1}{c|}{Evaluation Strategy}         & \multicolumn{1}{c}{steps}                           \\
Optimizer                   & Adamw                           \\
\rowcolor{gray!20} 
\multicolumn{1}{c|}{GPU}                         & \multicolumn{1}{c}{two NVIDIA A800-SXM4 (80G) GPUs} \\ \bottomrule[1.5pt]
\end{tabular}%
}
\vspace{-0.2cm}
\caption{Experimental hyperparameter settings}
\vspace{-0.5cm}
\label{tab: setting-exp}
\end{table}

\begin{table}[]
\resizebox{\columnwidth}{!}{%
\begin{tabular}{@{}c|c|c@{}}
\toprule[1.5pt]
\textbf{Method}                    & \textbf{Hyperparameter}                    & \textbf{Setting}                                                     \\ \midrule\midrule
\rowcolor{gray!20}
\multicolumn{1}{c|}{Prompt Tuning}             & \multicolumn{1}{c|}{prompt\_tuning\_init\_text}        & \multicolumn{1}{c}{Answer the following question as required.\textbackslash{}n} \\ \midrule
\multirow{4}{*}{P-Tuning} & num\_virtual\_tokens              & 20                                                          \\
                          & encoder\_hidden\_size             & 256                                                         \\
                          & encoder\_num\_layers              & 2                                                           \\
                          & encoder\_reparameterization\_type & MLP                                                         \\ \midrule
\rowcolor{gray!20}
\multicolumn{1}{c|}{IA3}                       & \multicolumn{1}{c|}{target\_modules}                   & \multicolumn{1}{c}{default}                                                     \\ \midrule
\multirow{3}{*}{LoRA}     & r                                 & 8,16,24,32,64                                               \\
                          & target\_modules                   & down\_proj, k\_proj, v\_proj, q\_proj, up\_proj, gate\_proj, o\_proj                                                         \\
                          & lora\_alpha                       & r $\times$ 2                                   \\ \midrule
\rowcolor{gray!20}
\multicolumn{1}{c|}{DoRA}                      & \multicolumn{1}{c|}{r}                                 & \multicolumn{1}{c}{8}                                                           \\ \midrule
PiSSA                     & r                                 & 8                                                           \\ \midrule
\rowcolor{gray!20}
\multicolumn{1}{c|}{HydraLoRA}                 & \multicolumn{1}{c|}{\#A | \#B}                           &\multicolumn{1}{c}{ 1 | 3,1 | 14}                                                    \\ \midrule
MOELoRA                   & \#A | \#B                           & 8 | 8                                                         \\ \midrule
\rowcolor{gray!20}
\multicolumn{1}{c|}{MTL-LoRA}                  & \multicolumn{1}{c|}{\#A | \#B}                           & \multicolumn{1}{c}{1 | 14}                                                        \\ \midrule
\multirow{2}{*}{MoLA}     & \#A | \#B                           & 8 | 8                                                         \\
                          & Type                              & Inverted-Triangle (MoLA-$\nabla$)                                  \\ \midrule
\multicolumn{1}{c}{\cellcolor{gray!20}{\multirow{2}{*}{CoLA}}}     & \cellcolor{gray!20}{r}                                 & \multicolumn{1}{c}{\cellcolor{gray!20}{8}}                                                           \\
     \multicolumn{1}{c}{\cellcolor{gray!20}{\multirow{-2}{*}{CoLA}}}                     & \cellcolor{gray!20}{\#A | \#B}                           & \multicolumn{1}{c}{\cellcolor{gray!20}{1 | 3,2 | 3,1 | 14, 4 | 10}}                                           \\ \bottomrule[1.5pt]
\end{tabular}%
}
\vspace{-0.2cm}
\caption{Hyperparameter settings used in our experiments for different PEFT methods.}
\vspace{-0.4cm}
\label{tab: setting-method}
\end{table}

Note that we present our experimental details, as shown in Tables~\ref{tab: setting-exp} and~\ref{tab: setting-method}.

\section{Additional Experimental Results}
\label{apx: results}

To verify the robustness of CoLA when faced with larger sample sizes (sample size = 1000, 2000, 3000, 4000, 5000, All), we selected PiSSA$_{r=16}$ and PiSSA$_{r=24}$ as the baselines for CoLA ($\text{\#}A=1, \text{\#}B=3$) and CoLA$^{\intercal}$ ($\text{\#}A=2, \text{\#}B=3$) in the generality domain, as shown in Table~\ref{tab:more}. From the table, it can be observed that as the sample size increases, CoLA and CoLA$^{\intercal}$ consistently outperform PiSSA${r=16}$ and PiSSA${r=24}$, which confirms the robustness of CoLA with respect to more samples. We also notice that having too many samples (ALL) does not necessarily lead to optimal performance, which may be due to the model's parameter scale not keeping up with the scaling law. 

\begin{table}[t]
\resizebox{\columnwidth}{!}{%
\begin{tabular}{@{}c|c|c|c|c|c|l@{}}
\toprule[1.5pt]
\textbf{Sample Size} & \multicolumn{1}{c|}{\textbf{1000}}      & \multicolumn{1}{c|}{\textbf{2000}}      & \multicolumn{1}{c|}{\textbf{3000}}      & \multicolumn{1}{c|}{\textbf{4000}}      & \multicolumn{1}{c|}{\textbf{5000}}      & \multicolumn{1}{c}{\textbf{ALL}}       \\ \midrule
\textbf{PiSSA$_{r=16}$}      & \multicolumn{1}{c|}{55.95}              & \multicolumn{1}{c|}{57.28}              & \multicolumn{1}{c|}{59.60}              & \multicolumn{1}{c|}{61.86}              & \multicolumn{1}{c|}{60.82}              & \multicolumn{1}{c}{57.37}              \\
\textbf{PiSSA$_{r=24}$}      & 55.62                                  & 52.02                                  & 60.88                                  & 61.00                                  & 60.85                                  & 57.29                                  \\
\textbf{CoLA}       & \multicolumn{1}{c|}{\cellcolor[HTML]{CBCEFB}\textbf{58.04}} & \cellcolor[HTML]{CBCEFB}\textbf{59.48} & \cellcolor[HTML]{CBCEFB}\textbf{62.49} & \cellcolor[HTML]{CBCEFB}\textbf{62.89} & \cellcolor[HTML]{CBCEFB}\textbf{62.14} & \multicolumn{1}{c}{\cellcolor[HTML]{CBCEFB}\textbf{59.53}} \\
\textbf{CoLA$^{\intercal}$}       & \multicolumn{1}{c|}{\cellcolor[HTML]{CBCEFB}\textbf{58.21}} & \cellcolor[HTML]{CBCEFB}\textbf{58.84} & \cellcolor[HTML]{CBCEFB}\textbf{61.54} & \cellcolor[HTML]{CBCEFB}\textbf{63.28} & \cellcolor[HTML]{CBCEFB}\textbf{63.33} & \multicolumn{1}{c}{\cellcolor[HTML]{CBCEFB}\textbf{60.56}} \\ \bottomrule[1.5pt]
\end{tabular}%
}
\vspace{-0.2cm}
\caption{Performance comparison of CoLA and PiSSA in the generity domain based on Llama-3.1-8B as the sample size increases.}
\label{tab:more}
\end{table}

\end{document}